\pdfoutput=1

\documentclass[11pt]{article}

\usepackage[final]{acl}

\usepackage{times}
\usepackage{latexsym}

\usepackage[T1]{fontenc}

\usepackage[utf8]{inputenc}
\usepackage{multirow}
\usepackage{longtable}
\usepackage{booktabs}
\usepackage{microtype}
\usepackage{amsmath}
\usepackage{amsfonts}
\usepackage{inconsolata}
\usepackage{makecell}
\usepackage{graphicx}
\usepackage{subcaption}
\title{Separate the Wheat from the Chaff: A Post-Hoc Approach to Safety Re-Alignment for Fine-Tuned Language Models}

\author{Di Wu, Xin Lu, Yanyan Zhao, Bing Qin\\
        Research Center for Social Computing and Information Retrieval \\ Harbin Institute of Technology, China\\
        \{dwu, xlu, yyzhao, qinb\}@ir.hit.edu.cn}

\begin{document}
\maketitle
\begin{abstract}
Although large language models (LLMs) achieve effective safety alignment at the time of release, they still face various safety challenges. A key issue is that fine-tuning often compromises the safety alignment of LLMs. To address this issue, we propose a method named \textbf{IRR} (\textbf{I}dentify, \textbf{R}emove, and \textbf{R}ecalibrate for Safety Realignment) that performs safety realignment for LLMs. The core of IRR is to identify and remove unsafe delta parameters from the fine-tuned models, while recalibrating the retained parameters. We evaluate the effectiveness of IRR across various datasets, including both full fine-tuning and LoRA methods. Our results demonstrate that IRR significantly enhances the safety performance of fine-tuned models on safety benchmarks, such as harmful queries and jailbreak attacks, while maintaining their performance on downstream tasks. The source code is available at: \url{https://github.com/pikepokenew/IRR}.

\end{abstract}

\section{Introduction}

In recent years, large language models (LLMs) have been widely used due to their significant success in various tasks \citep{qin2023chatgpt, zhao2023chatgpt}. A common paradigm for LLMs is “release and fine-tuning.” Before release, developers conduct safety alignment to achieve a safety-aligned model \citep{ouyang2022training}. After release, these LLMs are made available through fine-tuning APIs or open-source platforms, enabling users to further fine-tune them for specific downstream tasks.

In the “release and fine-tuning” paradigm, the parameters of LLMs change, and these changes are known as delta parameters, which improve performance on downstream tasks. However, this process often compromises the safety mechanisms established during safety alignment, reducing their value as reliable AI services. Specifically, training data that mixes harmful data with benign data, or even consists entirely of benign data, can significantly compromise the safety alignment of LLMs \citep{bhardwaj2023language, qi2023fine, wan2023poisoning, zhan2024removing, huang2024harmful}. Given the widespread use of the “release and fine-tuning” paradigm and its associated risks, a key objective is to ensure the safety realignment of fine-tuned models while maintaining their performance on downstream tasks.

\begin{figure}[!t]
    \centering
    \includegraphics[width=\linewidth]{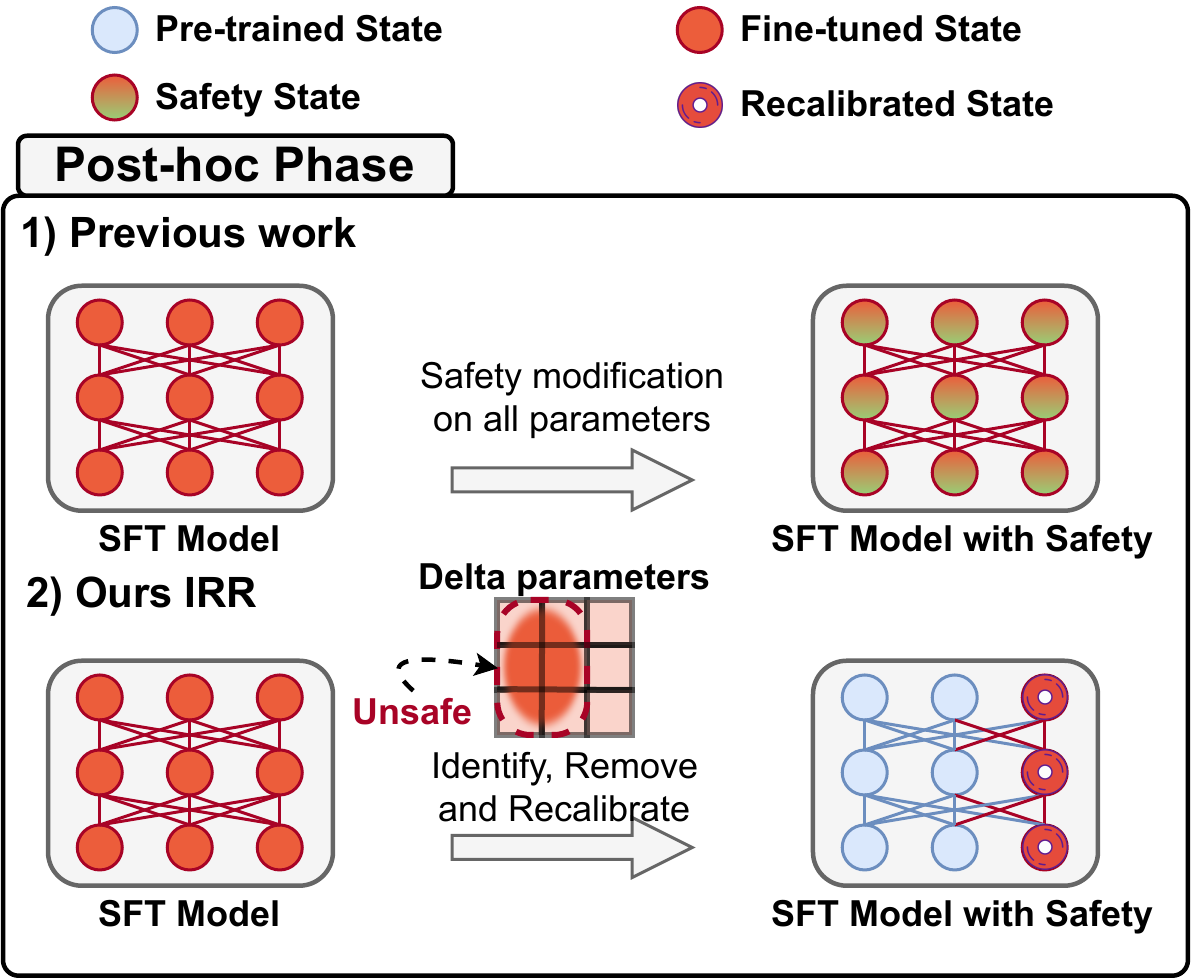}
    \caption{The illustration presents post-hoc approaches for safety realignment. Our method, IRR, first identifies and removes unsafe delta parameters, then recalibrates the remaining ones.}
    \label{fig:introduction}
\end{figure}

Recent works show directly modifying the parameters of the fine-tuned model can effectively and flexibly achieve this objective. A notable example is RESTA \citep{bhardwaj2024language}, which improves safety by merging a safety vector into all parameters (Figure \ref{fig:introduction}). However, current methods overlook the safety relevance and task relevance of parameters \citep{frankle2018lottery, panigrahi2023task, weiassessing}, as well as the trade-offs between them. As a result, applying the same safety modifications to all parameters without distinction may unintentionally harm those critical for downstream task performance. Additionally, some research shows that removing redundant delta parameters from fine-tuned models can restore the abilities that were forgotten due to fine-tuning \citep{panigrahi2023task, he2024localize, zhu2024model}. Inspired by this, we suggest separating unsafe delta parameters from the fine-tuned models to restore safety and recalibrating the retained parameters to minimize the performance loss, as shown in Figure \ref{fig:introduction}. This method can achieve a better balance between safety and downstream task performance.

Based on these insights, we propose \textbf{IRR} (\textbf{I}dentify, \textbf{R}emove, and \textbf{R}ecalibrate for Safety Realignment), a simple post-hoc method for the safety realignment of fine-tuned models. Specifically, the IRR method involves three steps: (1) To identify unsafe delta parameters in the fine-tuned model, we introduce a safety vector. This vector represents the changes that move the model from an unsafe state to a safe state. If the sign of a delta parameter disagrees with the safety vector, it can interfere with the model safety. We identify the delta parameters that cause safety interference and are important for safety as unsafe. (2) We remove the unsafe delta parameters from the fine-tuned models. (3) Unsafe delta parameters can compromise safety but may be important for downstream task performance. Removing them might reduce performance. To mitigate this, we recalibrate the retained parameters using precise weight compensation based on the inverse of the Hessian matrix.

We conducted extensive experiments to evaluate IRR under full fine-tuning and LoRA fine-tuning across various datasets. Compared to the baselines, IRR significantly improves model safety while preserving downstream task performance, achieving a Pareto improvement.

Our contributions can be summarized as follows:

\begin{itemize} 
\item We propose IRR, a novel safety realignment method that improves safety through three steps: identify, remove, and recalibrate.
\item IRR introduces a novel perspective on safety realignment, showing that combining safety interference and safety importance scores can effectively separate unsafe delta parameters from fine-tuned models to enhance safety.
\item Extensive experiments across various datasets, fine-tuning methods, and models show that IRR effectively restores safety while preserving downstream task performance, achieving Pareto improvements.
\end{itemize}

\begin{figure*}[!ht]
    \centering
    \includegraphics[width=\textwidth]{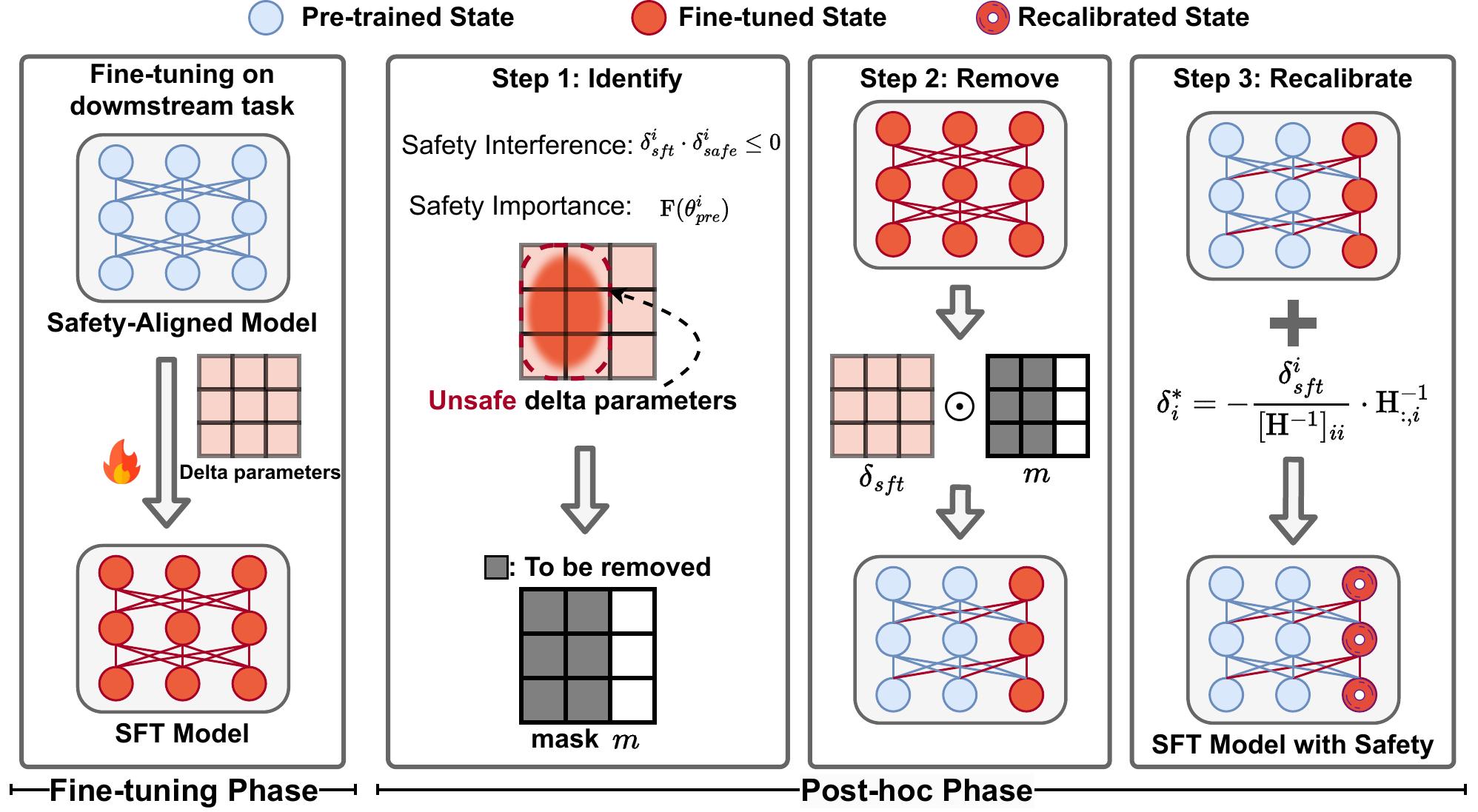}
    \caption{During fine-tuning phase, safety-aligned models acquire delta parameters that enhance downstream task performance, but these parameters may compromise model safety. In the post-hoc phase, IRR carefully identifies and removes unsafe delta parameters. It then computes compensatory values and adds them to the retained parameters, effectively restoring safety while preserving the model performance on downstream tasks.}
    \label{fig:methodology}
\end{figure*}

\section{Related Work}
\paragraph{LLMs Safety}
The safety of LLMs aims to mitigate potential safety risks arising from misuse or malicious use. Recent studies have identified vulnerabilities in the safety alignment of LLMs. \citet{yang2023shadow,bhardwaj2023language,zhan2024removing,huang2024harmful} demonstrated that even fine-tuning on small amounts of harmful data can significantly impact the safety of LLMs. \citet{qi2023fine} used more practical datasets, such as identity shift data and benign data like Alpaca, to undermine the safety alignment of LLMs. 

To address the safety compromises introduced by fine-tuning, current methods focus on four main phases: (1) Pre-processing phase, where \citet{zhao2023learning} uses catastrophic forgetting to filter harmful data; (2) Fine-tuning phase, where \citet{huang2024lazy} limits parameter updates to reduce safety loss; (3) Post-hoc phase after fine-tuning, where \citet{bhardwaj2024language} employs a merging safety vector approach to enhance safety, and \citet{zhao2024towards} introduces a patch to the safety vector to mitigate over-safety issues. Additionally, \citet{hsu2024safe} proposes the Safe LoRA, and \citet{yi2024safety} trains a safety sub-network within the fine-tuned model. \citet{huang2024antidote} employs model pruning to remove harmful parameters from the model. These methods do not require intervention during the inference time; (4) Inference time phase, where \citet{hazra-etal-2024-safety} removed harmful vectors and adjusted the latent space, while \citet{xu-etal-2024-safedecoding} improved model safety by increasing the probability of safety tokens during the decoding.

We focus on the post-hoc phase that does not require additional fine-tuning, thereby reducing computational costs while allowing flexible trade-offs between safety and downstream task performance.

\paragraph{Supervised Fine-Tuning and Delta Parameters.} Supervised fine-tuning (SFT) is a widely used method to enhance the performance of pre-trained LLMs on specific downstream tasks. This process involves changing the model parameters to improve task performance, with these alterations referred to as delta parameters. Recent studies have highlighted redundancy in these delta parameters in fine-tuned models. \citet{panigrahi2023task} addressed this issue by employing sub-network search to selectively prune delta parameters, retaining only a minimal subset necessary to achieve performance comparable to standard SFT. Similarly, \citet{yu2024language} introduced the DARE method, which involves randomly dropping a certain proportion of delta parameters and rescaling the remaining ones. \citet{zhu2024model} proposed the use of significance and sensitivity metrics to identify critical delta parameters. Our method also focuses on delta parameters, but emphasizes their role in balancing safety and downstream performance, rather than focusing solely on downstream tasks.

\paragraph{Model Pruning Technique}
As neural network models grow in size, model pruning techniques have been widely adopted to reduce computational costs \citep{cheng2017survey, liang2021pruning}. The goal of model pruning is to remove unnecessary parameters while maintaining model performance \citep{zhu2017prune}. The key to successful pruning is to assess the importance of parameters. For example, \citet{liu2021group} used the Fisher matrix to compute parameter importance and removed those. \citet{frantar2023sparsegpt} introduced SparseGPT, forming importance scores by solving a layer-wise reconstruction problem, while \citet{sunsimple} proposed the Wanda score, which assesses parameters using joint weight/activation metrics. Although our approach involves parameter removal similar to model pruning, we specifically focus on removing delta parameters rather than entire parameters.

\section{Approach}
We propose IRR, a novel method for safety realignment of fine-tuned models, which effectively restores safety while maintaining downstream task performance. The overall framework of IRR is illustrated in Figure \ref{fig:methodology}, and consists of the following three steps: (1) \textbf{Identify the Unsafe Delta Parameters}. This step identifies delta parameters that interfere with important parameters for safety alignment and marks these interfering delta parameters as unsafe. (2) \textbf{Remove the Unsafe Delta Parameters}. These unsafe delta parameters are removed and reverted to their original safe pre-trained states, improving the safety of fine-tuned models. (3) \textbf{Recalibrate the Retained Parameters}. Since some unsafe delta parameters may significantly affect downstream task performance, removing them could degrade performance. To mitigate this, we compute compensatory values and add them to the retained parameters.

\subsection{Identify the Unsafe Delta Parameters} In this step, we propose two strategies: \textbf{Safety Interference} and \textbf{Safety Importance} to identify the unsafe delta parameters. These strategies help separate unsafe delta parameters from fine-tuned models, thereby restoring safety.

\paragraph{Safety Interference} 
To identify unsafe delta parameters, it is crucial to clarify the direction of safe parameter updates. Therefore, we define a safety vector \(\delta_{safe}\), which represents the parameter differences when moving from the unaligned model to the safety-aligned model:
\begin{equation}
    \delta_\text{safe} = \theta_\text{align} - \theta_\text{unalign}
\end{equation}
Inspired by the concept of interference \citep{yadav2024ties}, we hypothesize that if a delta parameter $\delta^{i}$ has a sign disagreement with the safety vector $\delta_{safe}^{i}$, it causes safety interference that compromises model safety. Therefore, it is essential to identify delta parameters in $\delta_{sft}$ that exhibit sign disagreement with $\delta_{safe}$, forming a candidate set $\mathcal{U}$ of safety interference delta parameters, as defined by the following formula:
\begin{equation}
   \mathcal{U} = \{ \delta_\text{sft}^{i} \in \delta_\text{sft} \,|\, \delta_\text{sft}^{i} \cdot \delta_\text{safe}^{i} \leq 0, \, \forall i \}
\end{equation}

\paragraph{Safety Importance} After identifying the candidate set of safety interference delta parameters \(\mathcal{U}\), the next step is to determine which of these parameters poses a threat to model safety. Building on previous work \citep{liu2021group, matena2022merging}, we introduce the Fisher matrix \citep{fisher1922mathematical, amari1996neural} as a safety importance score to evaluate the significance of each parameter relative to the safety alignment of the original model.

To simplify computation for LLMs, we approximate the Fisher matrix by averaging the gradients of $N$ samples to estimate its diagonal \citep{kirkpatrick2017overcoming}. Our estimation is as follows:
\begin{equation}
    \hat{F}_\theta = \frac{1}{N} \sum_{i=1}^{N} \mathop{\mathbb{E}}\limits_{y \sim p_\theta(y|x_i)} \left( \nabla_\theta \log p_\theta(y|x_i) \right)^{2}
\end{equation}
where \( x_1, \ldots, x_N \) represent harmful queries, and the expectation over \( y \) indicates a safe refusal response that refuses harmful queries. Notably, the Fisher matrix is computed on the original model before fine-tuning and can be reused in the post-hoc phase without repeated computation.

Parameters with high safety importance scores are critical for the safety alignment of the original model. If delta parameters that cause safety interference exist on high-importance parameters, they may compromise the safety alignment of the model. Such delta parameters should be considered unsafe.

To identify the unsafe delta parameters, we extract the parameters in the top \(\rho\%\) based on their safety importance scores from the set \(\mathcal{U}\) and apply a mask to designate these delta parameters as unsafe. Additionally, \(s^{\prime}\) denotes the score of the parameter at the \(\rho\%\) position within \(\mathcal{U}\). The method for determining the final mask \(m\) is defined as follows:
\begin{equation}
m_i =
\begin{cases} 
1, & \text{if } \delta_\text{sft}^{i} \in \mathcal{U} \text{ and } s_i \geq s' \\ 
0, & \text{otherwise}
\end{cases} .
\end{equation}
Finally, we identify the delta parameters with a mask value of 1 as unsafe.

\subsection{Remove the Unsafe Delta Parameters}

In this step, we remove the identified unsafe delta parameters while retaining the remaining ones. We define the removing process. For the delta parameters \(\delta_\text{sft}\), we introduce a mask \(m \in \{0, 1\}^{|\theta|}\) to indicate which delta parameters are unsafe and will be removed. Meanwhile, \(\theta_{pre}\) denotes the parameters of the pre-trained safety-aligned model. The parameters of the model are computed as follows:
\begin{equation}
\widetilde{\theta}_\text{sft} = (1 - m) \odot \delta_\text{sft} + \theta_\text{pre} .
\end{equation}

\subsection{Recalibrate the Retained Parameters}
Removing unsafe delta parameters improves model safety but may degrade downstream performance, as some unsafe parameters are critical for tasks. To address this, we add compensatory values \(\delta_{sft}^{*}\) to the retained delta parameters \(\widetilde{\delta}_{sft}\), identified by the mask \(m\). During this step, these retained parameters are recalibrated to maintain task performance.

\begin{equation}
   \hat{\theta}_\text{sft} = \{ \widetilde{\theta}_\text{sft}^{i} + \delta_{i}^{*} \,|\, m_{i} = 0, \widetilde{\theta}_\text{sft}^{i} \in \widetilde{\theta}_\text{sft} \}
\end{equation}
Previous work on the Optimal Brain Surgeon (OBS) theory \citep{lecun1989optimal, hassibi1993optimal, zhu2024model} analyzed the change in loss caused by parameter alterations and studied the minimal perturbation required for the remaining parameters to minimize the loss. Based on these theories, our method applies compensatory values \(\delta_{i}^{*}\) to the retained delta parameters \(\theta_\text{sft}^{i}\), ensuring optimal performance on downstream tasks. The compensatory values for the retained parameters are computed using the following formula:
\begin{equation}
    \delta_{i}^{*} = -\frac{\theta_\text{sft}^{i} - \theta_\text{pre}^{i}}{[\mathrm{H}^{-1}]_{ii}} \cdot \mathrm{H}^{-1}_{:,i} ,
\end{equation}
Here, \(\mathrm{H}^{-1}\) represents the inverse of the Hessian matrix, and \(\mathrm{H}^{-1}_{:,m}\) denotes the \(m\)-th column of \(\mathrm{H}^{-1}\).

The identify, remove, and recalibrate steps are executed iteratively on the parameter matrix using a specified block size, continuing this process until the entire parameter matrix has been traversed.

\section{Experimental Setup}
We conducted experiments using both full fine-tuning and LoRA \citep{hulora}. The results for full fine-tuning are presented in the main text, while LoRA results are in the Appendix \ref{sec:appendix_lora_experimental_details}.

\paragraph{Model} Our experiments are conducted on the widely used open-source model Llama-2-7b-chat \citep{touvron2023llama}, which has been fine-tuned to follow instructions, align with human preferences, and ensure strong safety. Additionally, we perform LoRA fine-tuning experiments based on the Llama-3-8B-Instruct \citep{dubey2024llama}. Supervised fine-tuning (SFT) is conducted using the LLaMA Factory \footnote{\url{https://github.com/hiyouga/LLaMA-Factory}} \citep{zheng-etal-2024-llamafactory}, and the resulting models are referred to as domain-specific fine-tuned models.

\paragraph{Dataset} For Llama-2, we utilized three datasets to obtain the SFT models: GSM8K \citep{cobbe2021training} for Math, CodeAlpaca-20k \footnote{\url{https://huggingface.co/datasets/sahil2801/CodeAlpaca-20k}} for Code, and Chinese Alpaca \citep{taori2023stanford} for Chinese capability. Following the setting of \citet{bhardwaj2024language}, we incorporated an additional 50K English instances into the Chinese Alpaca dataset to ensure the ability of the model to respond to English instructions. For Llama-3, we use the MathInstruct \citep{yuemammoth} dataset to obtain the SFT model.

\paragraph{Baselines} IRR and IRR$_d$ refer to the application of the method on SFT models without and with DARE \citep{yu2024language}, respectively. We compared the IRR and IRR$_d$ method against several baselines:
\begin{itemize}
    \item \textbf{SFT} involves fine-tuning on downstream task data using a language modeling objective.
    \item \textbf{DARE} \citep{yu2024language} applies a drop-and-rescale operation on the delta parameters of the SFT model.
    \item \textbf{Safe LoRA} \citep{hsu2024safe} maps the delta parameter matrix of the fine-tuned model into the subspace of safe vectors, resulting in a more secure fine-tuned model.
    \item \textbf{SafeDecoding} \citep{xu-etal-2024-safedecoding} identifies and amplifies the probabilities of safe tokens in generated content while reducing the probabilities of unsafe tokens, thereby enhancing model safety.
    \item \textbf{RESTA} \citep{bhardwaj2024language} improves the safety of fine-tuned model by incorporating safety vectors. Specifically, \textbf{RESTA} and \textbf{RESTA}$_d$ refer to methods that integrate safety vectors into SFT models without and with DARE \citep{yu2024language}, respectively.
\end{itemize}

\paragraph{Computing Safety Vectors and Fisher Matrix} According to \citet{bhardwaj2024language}, we define the safety vector \(\delta_{safe}\) as the difference in parameters between the aligned and unaligned models. The unaligned model is fine-tuned using a harmful question-answer dataset. We extracted 1,000 labeled harmful question-answer pairs from the BeaverTails dataset \citep{ji2024beavertails} for training. Safe LoRA and RESTA use the same safety vector.

To compute the Fisher matrix, we relied on the same set of harmful questions but generated safe responses using the aligned model to create a safety dataset. This safety dataset serves as the calibration dataset for computing the Fisher matrix.




\begin{table*}[!ht]
\centering
\LARGE
\begin{subtable}[t]{\textwidth}
\label{tab:fft_gsm8k}
\resizebox{\textwidth}{!}{%
\begin{tabular}{@{}lccccc|ccccccc|c@{}}
\toprule
\multirow{2}{*}{\textbf{Method}} & \multicolumn{5}{c|}{\textbf{Safety Score}} & \multicolumn{7}{c|}{\textbf{Jailbreak Safety Score}} & \multirow{2}{*}{\textbf{ \makecell{Math$\uparrow$ \\ (GSM8K)} }} \\ 
  & \textbf{CATQA} & \textbf{HEx-PHI} & \textbf{Salad-Base} & \textbf{Avg} & \textbf{$\Delta$}$\downarrow$ & \textbf{GPTFuzz} & \textbf{TAP} & \textbf{GCG} & \textbf{AutoDAN} & \textbf{Template} & \textbf{Avg} & \textbf{$\Delta$}$\downarrow$ \\ \midrule
\textbf{SFT} 
    & 78.85 & 71.52 & 90.01 & 80.12 & 19.83 & 32.11 & 44.76 & 23.91 & 36.96 & 38.52 & 35.25 & 47.71 & \textbf{43.06} \\
\textbf{DARE} 
    & 78.97 & 71.21 & 90.43 & 80.20 & 19.75 & 32.51 & 41.43 & 23.91 & 36.10 & 38.58 & 34.51 & 48.05 & \underline{42.99} \\
\textbf{SafeDecoding} 
    & 93.52  & 95.45  & 96.90  & 95.29  & 4.66 & 78.39 & 85.24 & 92.12 & 52.44 & 52.44 & 76.35 & 6.05 & 38.67 \\
\textbf{Safe LoRA} 
    & 99.88 & 100.00  & 99.86  & \textbf{99.91} & \textbf{0.04} & 74.73 & 96.19 & 89.13 & 65.04 & 89.58 & \textbf{82.93} & \textbf{0.03} & 22.61 \\
\textbf{RESTA} 
    & 99.33  & 99.09  & 99.34  & 99.26  & 0.69 & 54.51 & 76.19 & 89.13 & 86.82 & 78.24 & 76.98 & 5.98 & 41.93 \\
\textbf{RESTA$_d$} 
    & 99.52  & 98.79  & 99.20  & 99.17  & 0.78 & 55.10 & 75.71 & 88.86 & 84.24 & 77.49 & 76.28 & 6.68 & 41.77 \\ 
\textbf{IRR}
    & 99.58  & 99.39  & 99.72  & 99.56  & 0.39 & 59.56 & 82.86 & 85.05 & 91.12 & 79.68 & 79.65 & 3.31 & 42.91 \\ 
\textbf{IRR$_d$}
    & 99.70  & 99.70  & 99.77 & \underline{99.72} & \underline{0.23} & 60.36 & 84.29 & 86.96 & 93.12  & 82.10 & \underline{82.10} & \underline{0.86} & 42.61 \\ 
\bottomrule
\end{tabular}%
}
\caption{The results of fine-tuning on GSM8K and performing safety realignment.}
\end{subtable}

\begin{subtable}[t]{\textwidth}
\resizebox{\textwidth}{!}{%
\begin{tabular}{@{}lccccc|ccccccc|c@{}}
\toprule
\multirow{2}{*}{\textbf{Method}} & \multicolumn{5}{c|}{\textbf{Safety Score}} & \multicolumn{7}{c|}{\textbf{Jailbreak Safety Score}} & \multirow{2}{*}{\textbf{ \makecell{Code$\uparrow$ \\ (HumanEval)} }} \\ 
  & \textbf{CATQA} & \textbf{HEx-PHI} & \textbf{Salad-Base} & \textbf{Avg} & \textbf{$\Delta$}$\downarrow$ & \textbf{GPTFuzz} & \textbf{TAP} & \textbf{GCG} & \textbf{AutoDAN} & \textbf{Template} & \textbf{Avg} & \textbf{$\Delta$}$\downarrow$ & \\ \midrule
\textbf{SFT} 
    & 64.48 & 62.42 & 83.07 & 69.99 & 29.96 & 24.98 & 40.00 & 17.39 & 25.79 & 29.04 & 27.44 & 55.92 & \textbf{19.02} \\
\textbf{DARE} 
    & 64.18 & 63.03 & 83.26 & 70.16 & 29.79 & 26.16 & 42.38 & 19.02 & 25.50 & 28.91 & 28.40 & 54.56 & 18.66 \\
\textbf{SafeDecoding} 
    & 92.73 & 86.06 & 96.72 & 91.83 & 8.12 & 37.66 & 65.71 & 36.68 & 28.37 & 47.83 & 43.25 & 39.71 & 17.62 \\
\textbf{Safe LoRA} 
    & 99.94  & 99.70  & 99.91  & \textbf{99.85} & \textbf{0.10} & 77.60 & 96.67 & 60.87 & 80.80 & 80.82 & 79.35 & 3.61 & 11.89 \\
\textbf{RESTA} 
    & 99.70 & 96.97 & 99.62 & 98.76 & 1.19 & 66.30 & 81.90 & 68.21 & 93.12 & 83.47 & 78.60 & 4.36 & 14.88 \\
\textbf{RESTA$_d$} 
    & 99.58 & 96.67 & 99.67 & 98.64 & 1.31 & 66.80 & 82.86 & 69.57 & 91.69 & 83.63 & 78.91 & 4.05 & 15.61 \\ 
\textbf{IRR}
    & 99.27 & 97.88 & 99.72 & 98.96 & 0.99 & 71.46 & 85.24 & 67.93 & 91.69 & 89.28 & \underline{81.12} & \underline{1.84} & \underline{18.96} \\ 
\textbf{IRR$_d$}
    & 99.58 & 98.48 & 99.81 & \underline{99.29} & \underline{0.66} & 75.12 & 87.62 & 71.20 & 93.12 & 90.59 & \textbf{83.53} & \textbf{-0.57} & \textbf{19.02} \\ 
\bottomrule
\end{tabular}%
}
\label{tab:fft_code}
\caption{The results of fine-tuning on CodeAlpaca-20k and performing safety realignment.}
\end{subtable}

\begin{subtable}[t]{\textwidth}
\resizebox{\textwidth}{!}{%
\begin{tabular}{@{}lccccc|ccccccc|c@{}}
\toprule
\multirow{2}{*}{\textbf{Method}} & \multicolumn{5}{c|}{\textbf{Safety Score}} & \multicolumn{7}{c|}{\textbf{Jailbreak Safety Score}} & \multirow{2}{*}{\textbf{ \makecell{Chinese$\uparrow$ \\ (MMMLU)} }} \\ 
  & \textbf{CATQA} & \textbf{HEx-PHI} & \textbf{Salad-Base} & \textbf{Avg} & \textbf{$\Delta$}$\downarrow$ & \textbf{GPTFuzz} & \textbf{TAP} & \textbf{GCG} & \textbf{AutoDAN} & \textbf{Template} & \textbf{Avg} & \textbf{$\Delta$}$\downarrow$ & \\ \midrule
\textbf{SFT} 
    & 89.09 & 85.76 & 95.31 & 90.05 & 0.90 & 66.20 & 46.19 & 34.78 & 64.47 & 69.42 & 56.21 & 26.75 & \underline{36.85} \\
\textbf{DARE} 
    & 88.79 & 85.45 & 95.22 & 89.82 & 10.13 & 66.20 & 48.57 & 34.24 & 65.33 & 69.88 & 56.84 & 26.12 & 36.78 \\
\textbf{SafeDecoding} 
    & 96.61 & 91.21 & 97.70 & 95.17 & 4.78 & 81.27 & 66.19 & 62.77 & 76.22 & 77.79 & 72.85 & 10.11 & 23.29 \\
\textbf{Safe LoRA} 
    & 99.88  & 100.00 & 99.86 & \textbf{99.91} & \textbf{0.04} & 74.73 & 96.19 & 89.13 & 65.04 & 89.58 & 82.93 & 0.03 & 22.61 \\
\textbf{RESTA} 
    & 98.91 & 98.18 & 99.39 & 98.83 & 1.12 & 91.28 & 78.57 & 79.89 & 98.85 & 96.01 & \textbf{88.92} & \textbf{-5.96} & 33.03 \\
\textbf{RESTA$_d$} 
    & 99.03 & 98.18 & 99.48 & 98.90 & 1.05 & 89.89 & 77.62 & 79.35 & 99.14 & 95.95 & \underline{88.39} & \underline{-5.43} & 32.40 \\ 
\textbf{IRR}
    & 98.91 & 98.18 & 99.39 & 98.83 & 1.12 & 91.58 & 78.10 & 62.23 & 99.43 & 95.23 & 85.31 & -2.35 & \textbf{37.08} \\ 
\textbf{IRR$_d$}
    & 98.85 & 99.39 & 99.58 & \underline{99.27} & \underline{0.68} & 92.17 & 81.90 & 62.23 & 100.00 & 95.95 & 86.45 & -3.49 & 36.82 \\ 
\bottomrule
\end{tabular}%
}
\label{tab:fft_alpaca_zh}
\caption{The results of fine-tuning on Alpaca Chinese and performing safety realignment.}
\end{subtable}

\caption{We evaluate safety using harmful benchmarks and jailbreak attacks. A higher \textbf{safety score} indicates better safety, while \(\Delta\) represents the difference in safety score compared to the original model. Higher performance in downstream tasks reflects better capability. The best and second-best results are highlighted in \textbf{bold} and \underline{underlined}.}

\label{tab:fft_experiment}

\end{table*}

\paragraph{Evaluation Setup} To comprehensively evaluate the safety and robustness of LLMs, we considered two evaluation setups: (1) Harmful query benchmark and (2) Jailbreak attacks.

For the harmful query benchmark, we used three datasets: 1) CATQA \citep{bhardwaj2024language}, a multilingual dataset with English, Chinese, and Vietnamese; 2) HEx-PHI \citep{qi2023fine}, which includes 330 harmful queries based on usage policies from Meta and OpenAI; 3) Salad-Base \citep{li-etal-2024-salad}, covering 6 domains, 16 tasks, and 66 categories. We did stratified sampling on 10\% of the Salad-Base, resulting in 2,132 harmful queries.

To assess robustness against jailbreak attacks, we used the Salad-Attack \citep{li-etal-2024-salad} dataset, which simulates various attack attempts using methods from GPTFuzzer \citep{yu2023gptfuzzer}, TAP \citep{mehrotra2023tree}, GCG \citep{zou2023universal}, AutoDAN \citep{liuautodan}, and human-designed templates, all derived from the Salad-Base dataset.

We evaluated the downstream task performance of the SFT model using GSM8K \citep{cobbe2021training}, HumanEval \citep{chen2021evaluating}, and the Chinese version of MMMLU \footnote{\url{https://github.com/openai/simple-evals}} \citep{hendrycksmeasuring}. For Llama3, we conducted evaluations on various mathematical tasks, with detailed settings provided in the Appendix \ref{sec:appendix_llama3_experimental_details}.

\paragraph{Evaluation Metrics} We utilize MD-Judge \citep{li-etal-2024-salad}, a content moderation model based on LLMs, to evaluate the harmfulness of question-answer pairs, including responses to harmful requests and jailbreak attacks. We report the safety of the model as \textbf{Safety Score}, defined as the proportion of responses assessed as harmless by MD-Judge to all annotated responses. A higher score indicates a safer model.

\begin{figure*}[!ht]
	\centering
	\begin{subfigure}{0.32\textwidth}
		\centering
		\includegraphics[width=1.0\textwidth]{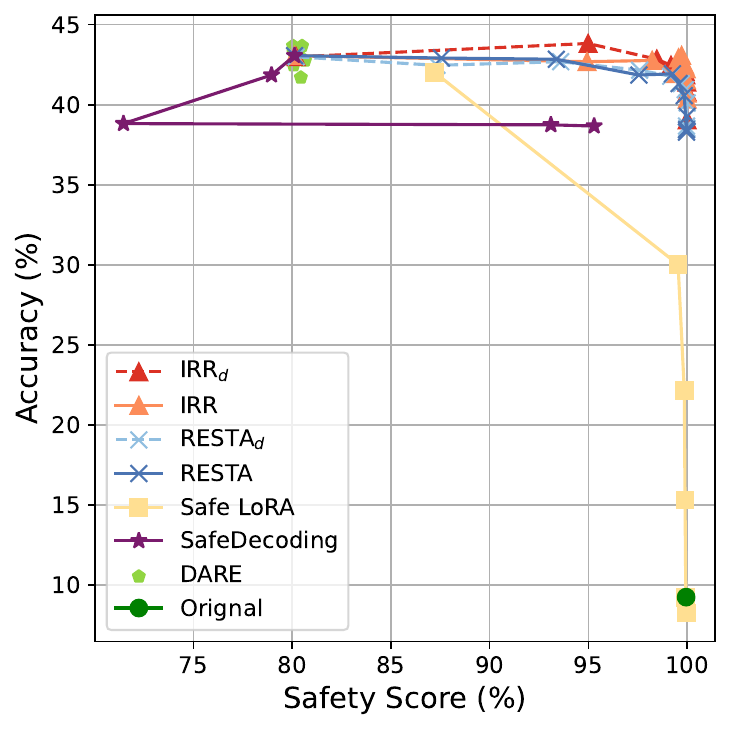}
		\caption{Math}
		\label{math_safety}
	\end{subfigure}
	\centering
	\begin{subfigure}{0.32\textwidth}
		\centering
		\includegraphics[width=1.0\textwidth]{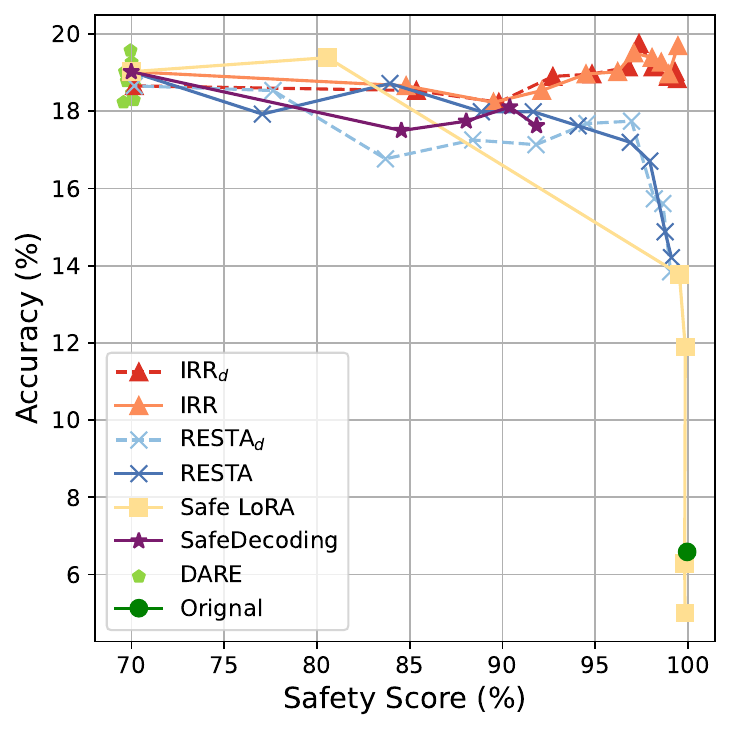}
		\caption{Code}
		\label{code_safety}
	\end{subfigure}
	\centering
	\begin{subfigure}{0.32\textwidth}
		\centering
		\includegraphics[width=1.0\textwidth]{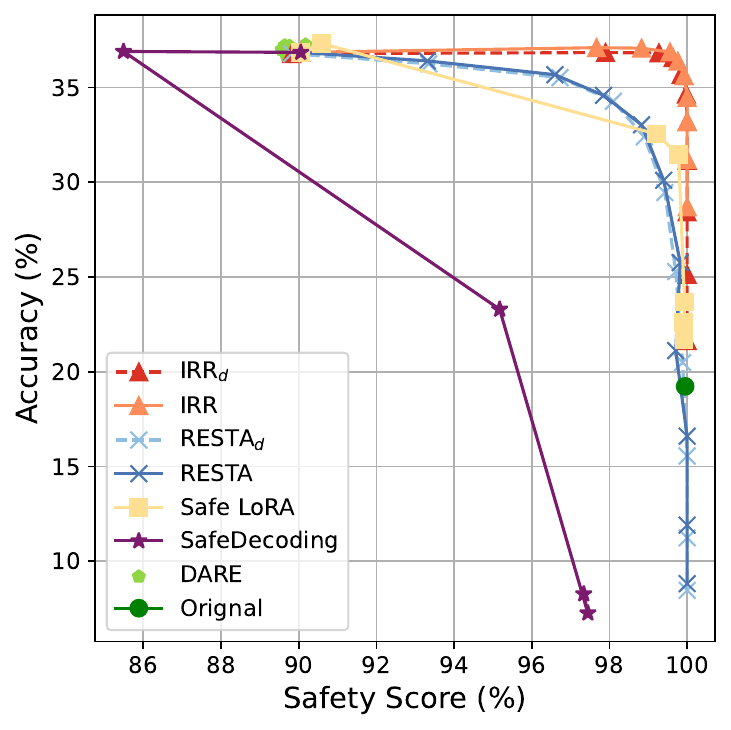}
		\caption{Chinese}
		\label{chinese_safety}
	\end{subfigure}
        \caption{We show the trend of “downstream task performance vs. safety score” based on the \textbf{Harmful Benchmark}. Our method, IRR, outperforms baseline methods, maintaining downstream task performance as safety improves.}

	\label{fig:safety_performance_pareto}
\end{figure*}

\begin{figure*}[!ht]
	\centering
	\begin{subfigure}{0.32\textwidth}
		\centering
		\includegraphics[width=1.0\textwidth]{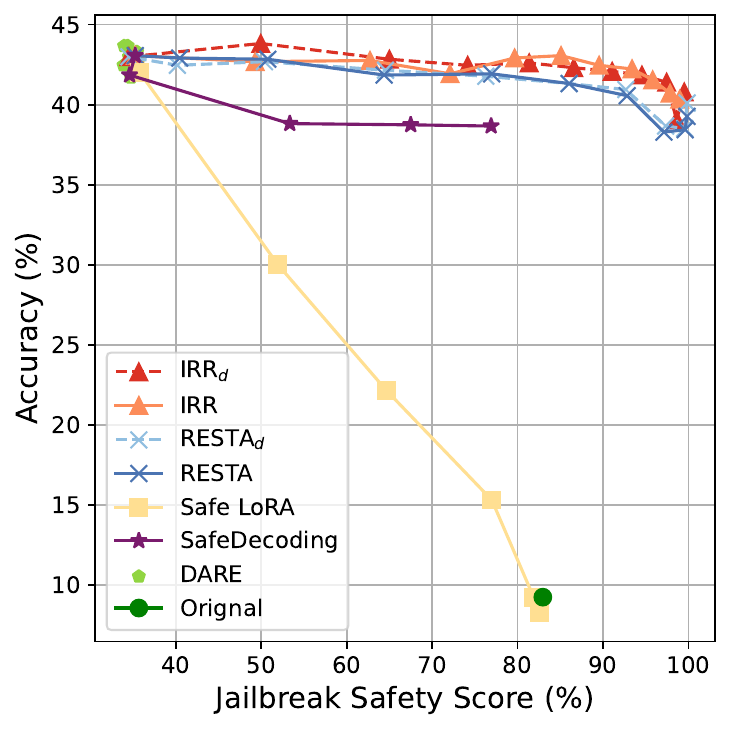}
		\caption{Math}
		\label{math_jailbreak}
	\end{subfigure}
	\centering
	\begin{subfigure}{0.32\textwidth}
		\centering
		\includegraphics[width=1.0\textwidth]{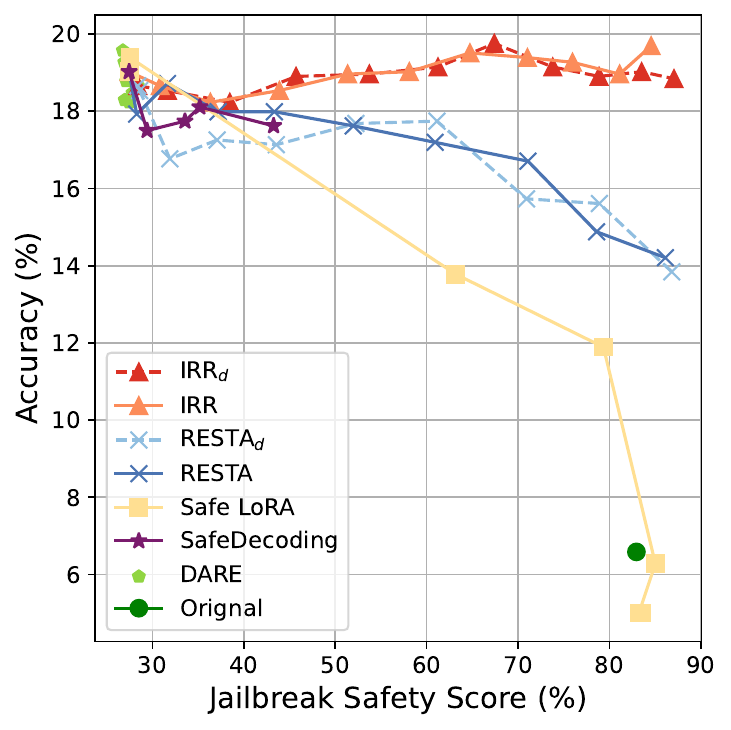}
		\caption{Code}
		\label{code_jailbreak}
	\end{subfigure}
	\centering
	\begin{subfigure}{0.32\textwidth}
		\centering
		\includegraphics[width=1.0\textwidth]{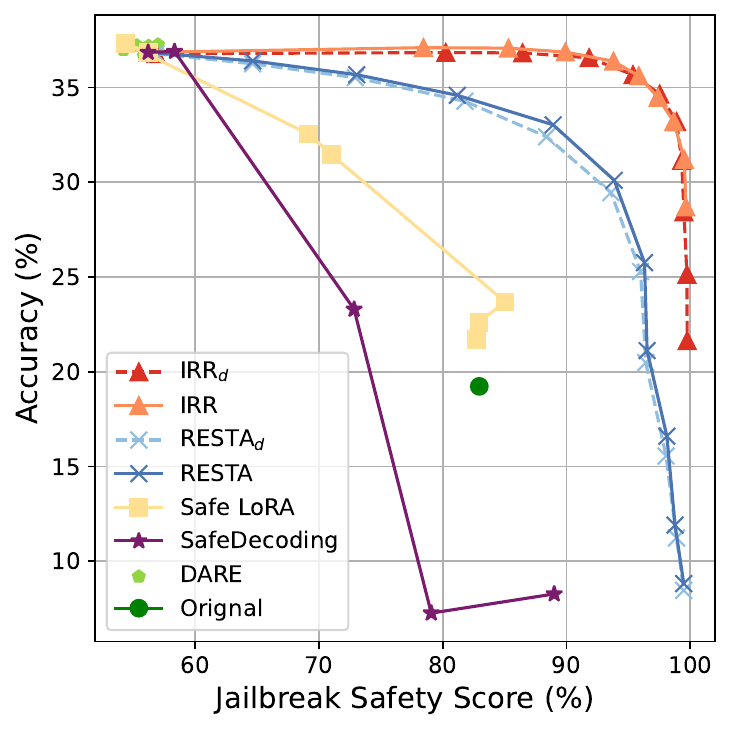}
		\caption{Chinese}
		\label{chinese_jailbreak}
	\end{subfigure}
	\caption{We show the trend of “downstream task performance vs. safety score” based on the \textbf{Jailbreak Attack}. Our method, IRR, outperforms baseline methods, maintaining downstream task performance as safety improves.}
	\label{fig:jailbreak_performance_pareto}
\end{figure*}

\section{Results and Discussions}

As shown in Table \ref{tab:fft_experiment}, all methods except DARE improved the safety of the SFT model. Although Safe LoRA enhanced model safety, it significantly reduced downstream task performance. For example, the accuracy on Math tasks dropped from 43.06 to 22.61. This indicates that projecting delta parameters into a safe subspace may disrupt parameters that are critical for downstream tasks.

The RESTA method also improved safety, with performance degradation varying across tasks. For example, the accuracy on Math tasks decreased by only 1.13, while accuracy on Code decreased by 4.14 and accuracy on Chinese decreased by 3.82. We also observed that the random dropout and scaling operations in the DARE method did not significantly improve safety, even when combined with RESTA or IRR, as seen in RESTA$_d$ and IRR$_d$.



In contrast, our IRR method maintains downstream task performance almost unchanged while enhancing safety compared to the SFT model.

\subsection{IRR Achieves Pareto Improvement} To investigate the trade-off between downstream task performance and safety during the safety improvement, we plotted the relationship between performance and safety (see Figure \ref{fig:safety_performance_pareto} and Figure \ref{fig:jailbreak_performance_pareto}). In the harmful benchmark and jailbreak attack, we observed that both RESTA and RESTA$_d$ maintained stable performance in the initial stages of safety improvements. However, as safety increased, their performance gradually declined, particularly for Code and Chinese tasks. In contrast, both IRR and IRR$_d$ consistently performed well at the safety frontier. Notably, even at safety levels close to those of the original model, IRR outperformed RESTA and RESTA$_d$ in downstream task performance, especially in jailbreak attacks. Additionally, the LoRA experiment results detailed in Appendix \ref{sec:appendix_lora_experimental_details} followed a similar trend as full fine-tuning, confirming the effectiveness of the IRR method.

\subsection{Ablation Study} 
We conducted an ablation study on the IRR method and reported the “downstream task performance vs. safety” trade-off curves on jailbreak attacks in Figure \ref{fig:ablation_analysis}. Additional ablation experiments can be found in the Appendix \ref{sec:appendix_ablations}.

\paragraph{Identifying the Unsafe Parameters.} We ablated the identification step (IRR w/o ID) and replaced it with a random selection of delta parameters to be removed. As shown in Figure \ref{fig:ablation_analysis}, skipping the identification step typically results in a significant drop in downstream task performance as a trade-off for improved safety. This highlights the critical importance of identifying unsafe delta parameters.

\paragraph{Safety Interference.} IRR uses safety interference together with the safety importance score to identify unsafe delta parameters. We ablated the safety interference strategy in the identification step (IRR w/o SI). As shown in Figure \ref{fig:ablation_analysis}, relying only on the safety importance score to identify unsafe delta parameters also leads to significant degradation in downstream task performance. This indicates that safety interference plays an important role.

\paragraph{Recalibration.} We ablated the recalibrate step (IRR w/o Recal). As shown in Figure \ref{fig:ablation_analysis}, removing recalibration resulted in performance degradation, although the impact was relatively minor.

These results validate the effectiveness of the IRR method.

\begin{figure*}[!ht]
	\centering
	\begin{subfigure}{0.32\textwidth}
		\centering
		\includegraphics[width=1.0\textwidth]{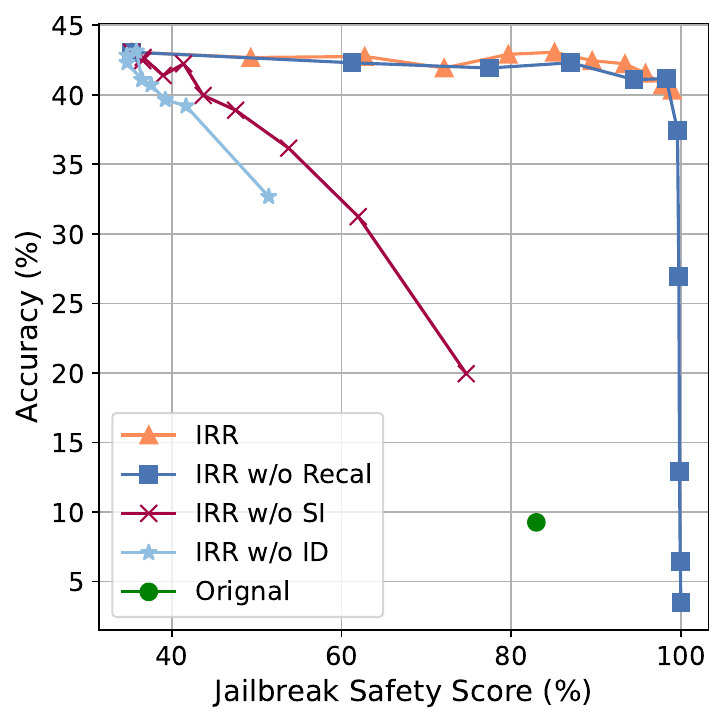}
		\caption{Math}
		\label{ablation_math_jailbreak}
	\end{subfigure}
	\centering
	\begin{subfigure}{0.32\textwidth}
		\centering
		\includegraphics[width=1.0\textwidth]{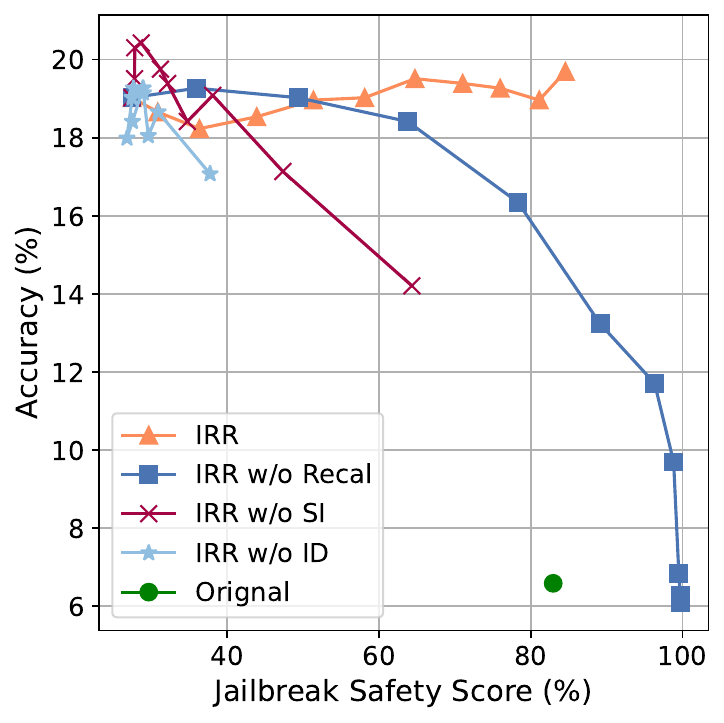}
		\caption{Code}
		\label{ablation_code_jailbreak}
	\end{subfigure}
	\centering
	\begin{subfigure}{0.32\textwidth}
		\centering
		\includegraphics[width=1.0\textwidth]{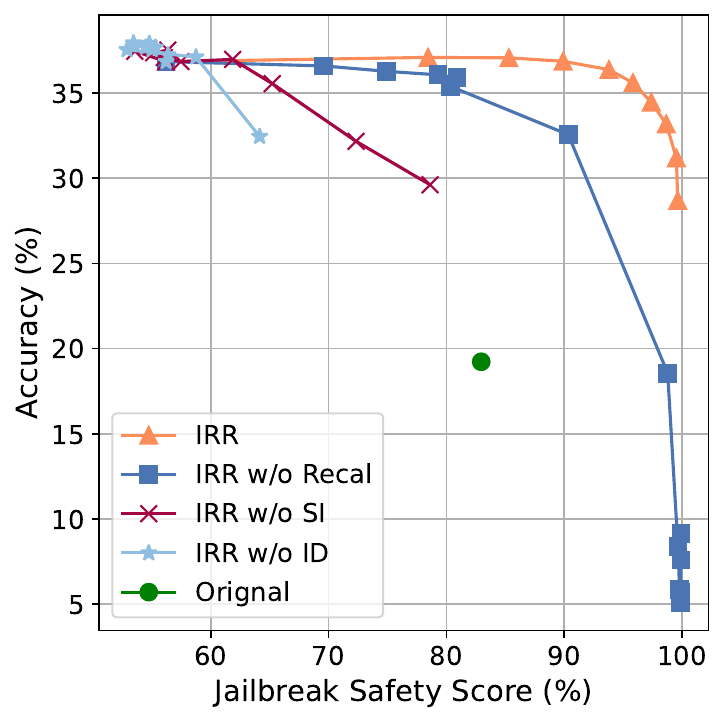}
		\caption{Chinese}
		\label{ablation_chinese_jailbreak}
	\end{subfigure}
	\caption{We present the results of the IRR ablation study using “downstream task performance vs. safety” curves.}
	\label{fig:ablation_analysis}
\end{figure*}

\begin{figure}[!t]
    \centering
    \begin{minipage}{0.7\linewidth}
        \centering
        \includegraphics[width=\linewidth]{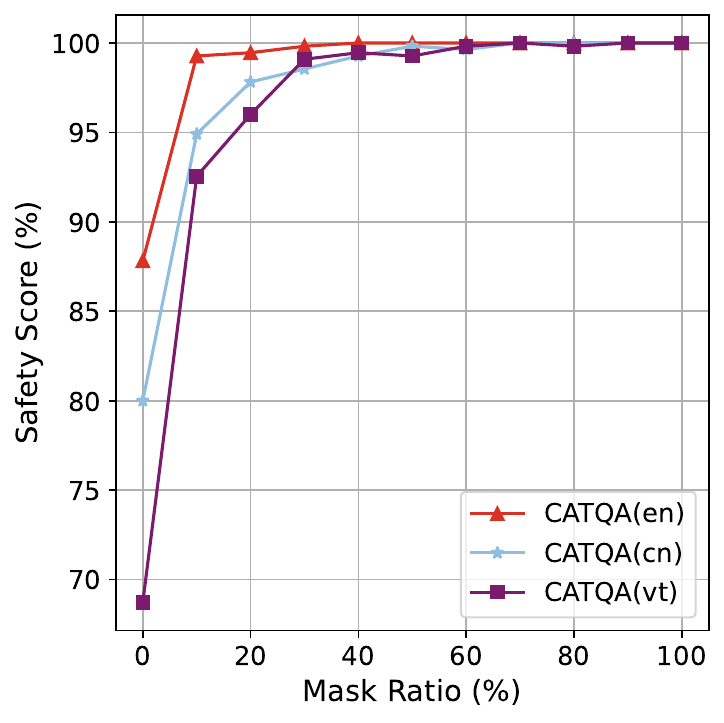}
    \end{minipage}\hfill
    \caption{We conduct a safety evaluation on the English, Chinese, and Vietnamese versions of CATQA. We compare the safety changes of harmful queries across different mask ratios and languages.}
    \label{fig:multilingual_catqa}
\end{figure}

\subsection{Cross-Language Safety Improvement} 
We assessed the safety of SFT models fine-tuned on mathematical datasets using the English, Chinese, and Vietnamese versions of the harmful benchmark CATQA (see Figure~\ref{fig:multilingual_catqa}). Additional experiments are provided in Appendix \ref{sec:appendix_multilingual}.

We observed that SFT models achieve lower safety scores in Chinese and Vietnamese compared to English. As the IRR mask ratio increases, the safety scores for harmful queries across different languages gradually improve, eventually approaching the safety level of the original model. Additionally, as shown in Appendix \ref{sec:appendix_multilingual}, different SFT models exhibit similar trends.

\begin{figure}[!ht]
    \centering
    \begin{minipage}{0.7\linewidth}
        \centering
        \includegraphics[width=\linewidth]{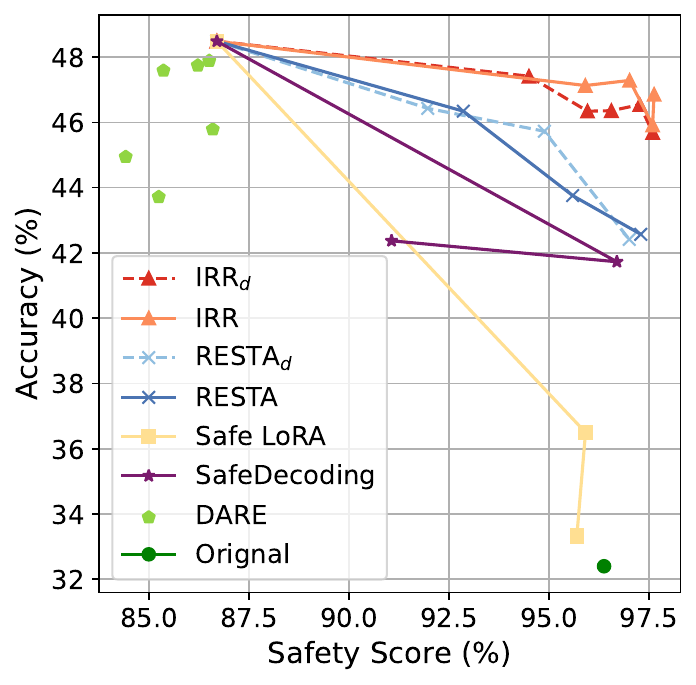}
    \end{minipage}\hfill
    \caption{We fine-tune the Llama-3-8B-Instruct model using LoRA on the MathInstruct dataset. We then evaluate the impact of IRR and baseline methods on the safety and mathematical capabilities of the SFT model.}
    \label{fig:llama3_mathinstruct_exp}
\end{figure}

\subsection{Efficacy of IRR Across Models} The IRR method is not restricted by any specific model architecture, allowing it to be applied across various models. To validate this claim, we conducted experiments by LoRA fine-tuning the LLama-3-8B-Instruct \citep{dubey2024llama}. The experimental results on harmful benchmarks are shown in Figure \ref{fig:llama3_mathinstruct_exp}. We evaluated the performance on several mathematical tasks and reported the average scores. The results demonstrate that for fine-tuned Llama3, the IRR method effectively improves model safety while maintaining downstream task performance, confirming its effectiveness. More detailed experimental results can be found in Appendix \ref{sec:appendix_llama3_experimental_details}.

\subsection{IRR with New Safety Vector}
\label{sec:IRR_new_safety_vector}
To study the effectiveness of IRR with different safety vectors, we sampled an additional 1,000 harmful question-answer pairs from Beavertails \citep{ji2024beavertails} and trained a new safety vector using LoRA. We present the results of IRR and RESTA with the new safety vector on models fine-tuned with LoRA on GSM8K \citep{cobbe2021training}.

\begin{figure}[!ht]
    \centering
    \begin{minipage}{0.7\linewidth}
        \centering
        \includegraphics[width=\linewidth]{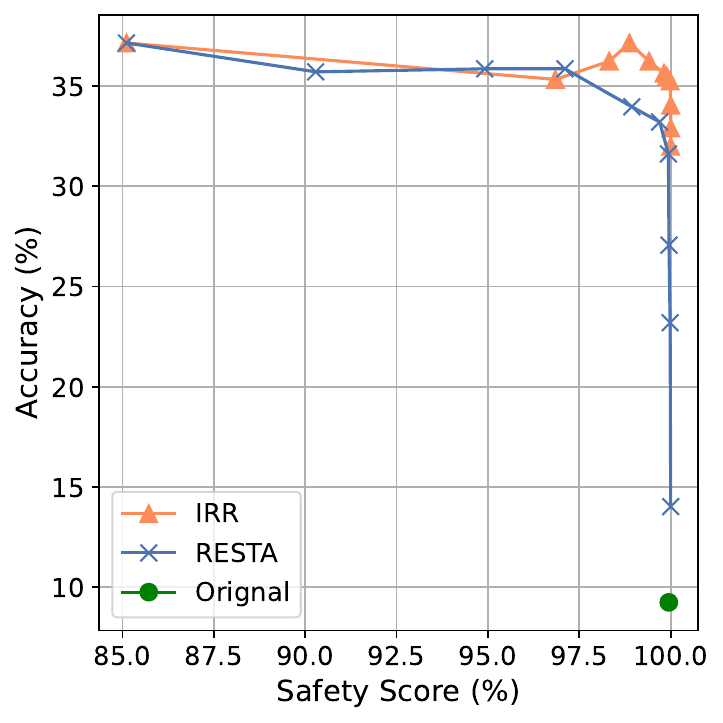}
    \end{minipage}\hfill
    \caption{Results of IRR using the new safety vector.}
    \label{fig:IRR_with_new_safety_vector}
\end{figure}

As shown in Figure \ref{fig:IRR_with_new_safety_vector}, IRR maintains better safety and downstream task performance than RESTA, even with the new safety vector. This indicates that the choice of safety vector does not affect IRR's effectiveness.

\subsection{IRR against Harmful Fine-tuning}
\label{sec:IRR_harmful_fine_tuning}
To further study IRR's effectiveness, we conducted harmful fine-tuning by mixing harmful data. We sampled 1,000 harmful question-answer pairs from Beavertails \citep{ji2024beavertails} and combined them with the GSM8K dataset for LoRA fine-tuning, keeping other settings unchanged. We also introduced a special IRR configuration, IRR$_{more}$, which removes additional delta parameters in ascending order of safety importance after removing all safety interference delta parameters.

\begin{figure}[!ht]
    \centering
    \begin{minipage}{0.7\linewidth}
        \centering
        \includegraphics[width=\linewidth]{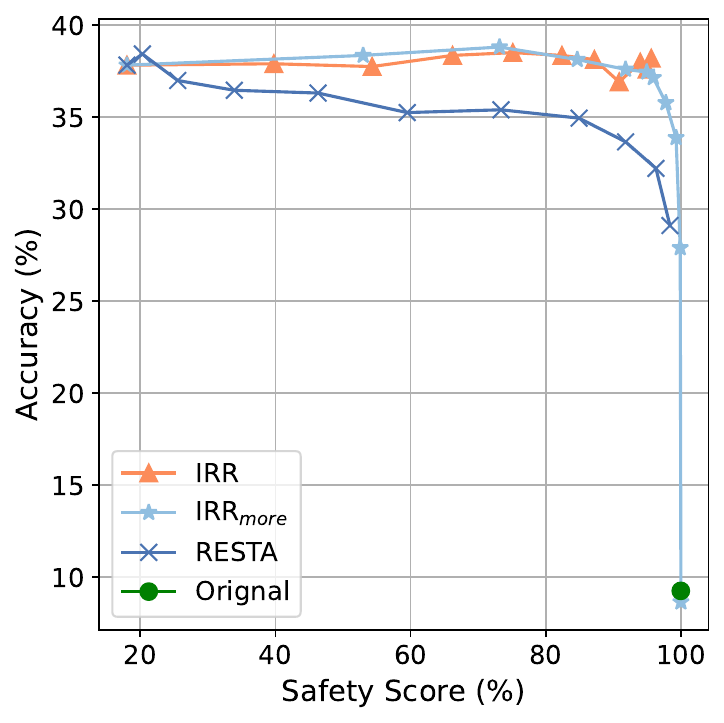}
    \end{minipage}\hfill
    \caption{Results of IRR against Harmful fine-tuning.}
    \label{fig:IRR_against_harmful_fine_tuning}
\end{figure}

The results from Figure \ref{fig:IRR_against_harmful_fine_tuning} show that, in the harmful fine-tuning setting, the IRR method improves safety while maintaining higher performance on downstream tasks compared to the RESTA method, demonstrating its effectiveness.

\section{Conclusion}
In this paper, we introduce a safety realignment method IRR, which enhances model safety by identifying and removing unsafe delta parameters while recalibrating the remaining ones. IRR significantly improves the effectiveness of safety realignment. Evaluations on various harmful query benchmarks and jailbreak attacks indicate that IRR considerably reduces the risks of fine-tuned models. Among various fine-tuning methods, datasets, and models, IRR outperforms baseline methods by improving model safety while maintaining downstream task performance, achieving Pareto improvements.

\section{Limitations}
Our work explores the important issue of safety realignment in fine-tuned models. While our findings offer valuable insights, they also highlight several limitations and directions for future research.

\paragraph{Multimodal Models.}
Due to budget constraints, we did not conduct experiments on multimodal models. However, we believe that safety assessments for images, speech, and other modalities could reveal more interesting insights, which we plan to consider in future work.

\paragraph{Our IRR.} 
Given the complex architecture of LLMs, our approach for obtaining safety vectors and evaluating the safety importance scores of parameters in the IRR method is relatively simple. Developing more robust and precise methods for these steps is necessary and should be a focus of future investigations.

Despite these limitations, we believe our work makes a new contribution to the field of safety alignment.

\section{Ethical consideration}
Ensuring ethical applications of artificial intelligence is critical. Our safety realignment method IRR enhances the safety of fine-tuned language models by reducing harmful content. The identification and removal operations effectively reduce harmful responses in fine-tuned models, while calibration ensures strong downstream task performance. Our framework demonstrates its effectiveness in improving the safety of fine-tuned models across different datasets. We advocate for ongoing collaboration among researchers, policymakers, and industry stakeholders to ensure that artificial intelligence development prioritizes human values, fairness, and safety. We remain committed to continuously evaluating and improving our approach to address ethical challenges.

\section{Potential Risks}
We now discuss the potential risks associated with our work. First, we highlight that the safety of fine-tuned models may be compromised, which could pose safety threats to users relying on these models for downstream tasks. We believe that improving safety will help the community benefit from advancements in secure large language models. 

On the other hand, our proposed safety re-alignment method may lead users to mistakenly believe that the resulting models are completely safe, which may not be the case. We only demonstrate improvements in safety based on the evaluations presented in this paper. This also poses potential safety risks to users. We recommend exercising caution when deploying language models and always conducting safety checks.

\section{Acknowledgements}
We thank the anonymous reviewers for their comments and suggestions. And we thank Weixiang Zhao, Bicheng Wang, Hao Yang for the suggestions on the writing of this work. This work was supported by the National Natural Science Foundation of China (NSFC) via grant 62441614 and 62176078.



\bibliography{anthology,custom}

\appendix

\section{Experimental Details}
\label{sec:appendix_experimental_details}
\paragraph{Hyperparameter Settings} For the Math, Code, and Chinese in Table \ref{tab:fft_experiment}, we set varying values for \(\rho\), specifically 60\%, 10\%, and 80\%, respectively. However, for each task, we can optimize the hyperparameters to achieve the best trade-off between downstream task performance and safety, as illustrated in Figure \ref{fig:safety_performance_pareto} and Figure \ref{fig:jailbreak_performance_pareto}. To plot the trade-off between safety and downstream task performance, the mask ratio starts at 10\% and increases in increments of 10\% until it reaches 100\%. The calculation of compensatory values in the recalibration step requires downstream task data. For this purpose, we extracted 1,000 samples from the corresponding downstream task dataset. We employed greedy decoding as our generation strategy. All experiments were conducted on 4 $\times$ A100 80GB GPUs.

We conducted full fine-tuning experiments on Llama-2-7B-chat \citep{touvron2023llama}, following the default configuration settings of Llama2. The initial learning rate was set to \(2.0 \times 10^{-5}\) and gradually decayed to zero using a cosine annealing schedule. The training batch size was set to 64. The number of epochs was set to 3, except for fine-tuning on the Alpaca Chinese dataset, for which only 1 epoch was used.

For experiments using Low-Rank Adaptation (LoRA) \citep{hulora} to fine-tune Llama-2-7B-chat \citep{touvron2023llama}, the query and value matrices in LoRA were adjusted with a rank of \(r=8\). We followed the default configuration settings of Llama2. The initial learning rate was set to \(2.5 \times 10^{-4}\) and gradually decayed to zero using a cosine annealing schedule. The training batch size was set to 64. The epoch was set to 5.

Both the training and inference of Llama-2 use the default system prompt and the chat template.

\begin{figure*}[t]
	\centering
	\begin{subfigure}{0.3\textwidth}
		\centering
		\includegraphics[width=1.0\textwidth]{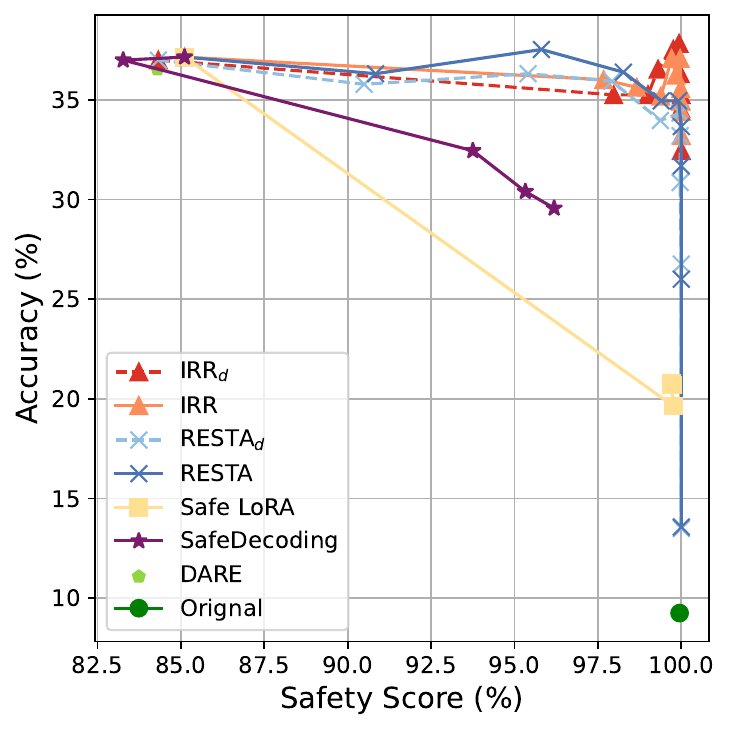}
		\caption{Math}
		\label{LoRA_math_safety}
	\end{subfigure}
	\centering
	\begin{subfigure}{0.3\textwidth}
		\centering
		\includegraphics[width=1.0\textwidth]{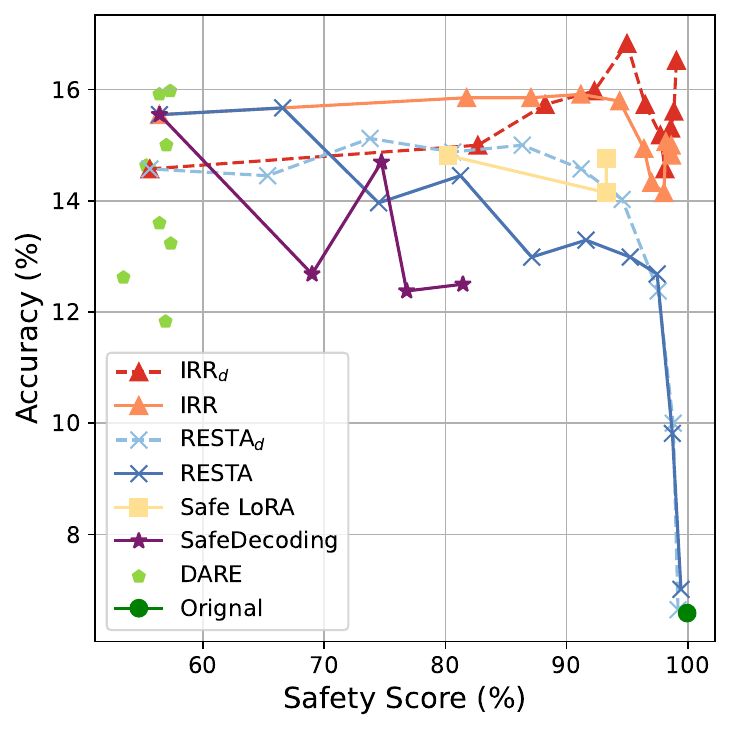}
		\caption{Code}
		\label{LoRA_code_safety}
	\end{subfigure}
	\centering
	\begin{subfigure}{0.3\textwidth}
		\centering
		\includegraphics[width=1.0\textwidth]{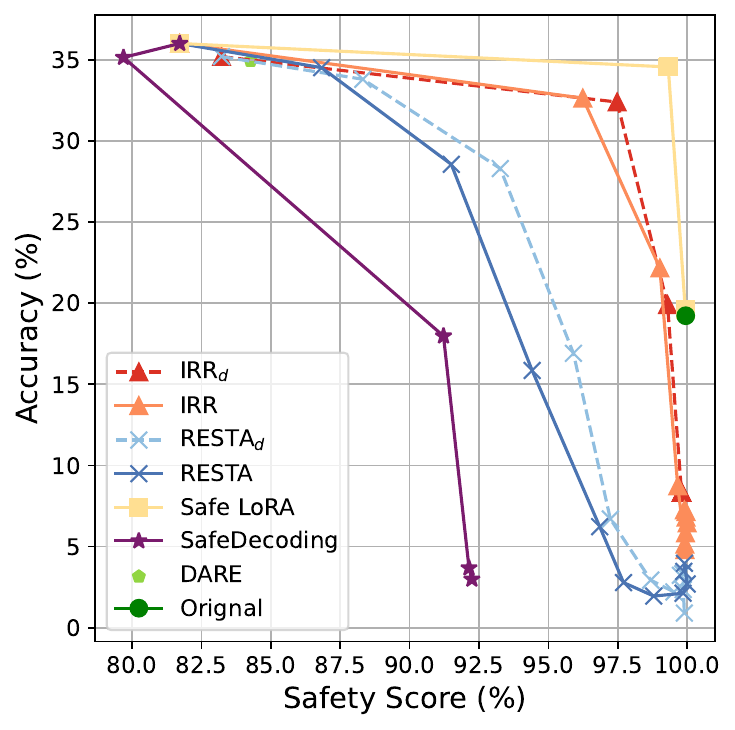}
		\caption{Chinese}
		\label{LoRA_chinese_safety}
	\end{subfigure}
	\caption{LoRA experimental results. We present the trends of "task performance versus safety score" on the \textbf{harmful benchmark}. Our method, IRR, outperforms baseline methods by improving safety while maintaining strong downstream task performance.}
	\label{fig:LoRA_safety_performance_pareto}
\end{figure*}

\begin{figure*}[t]
	\centering
	\begin{subfigure}{0.3\textwidth}
		\centering
		\includegraphics[width=1.0\textwidth]{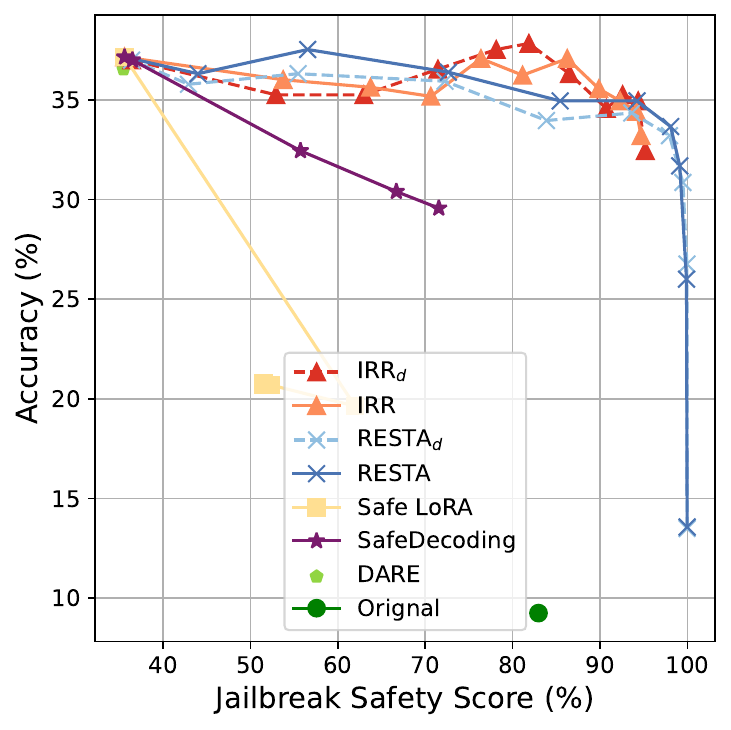}
		\caption{Math}
		\label{LoRA_math_jailbreak}
	\end{subfigure}
	\centering
	\begin{subfigure}{0.3\textwidth}
		\centering
		\includegraphics[width=1.0\textwidth]{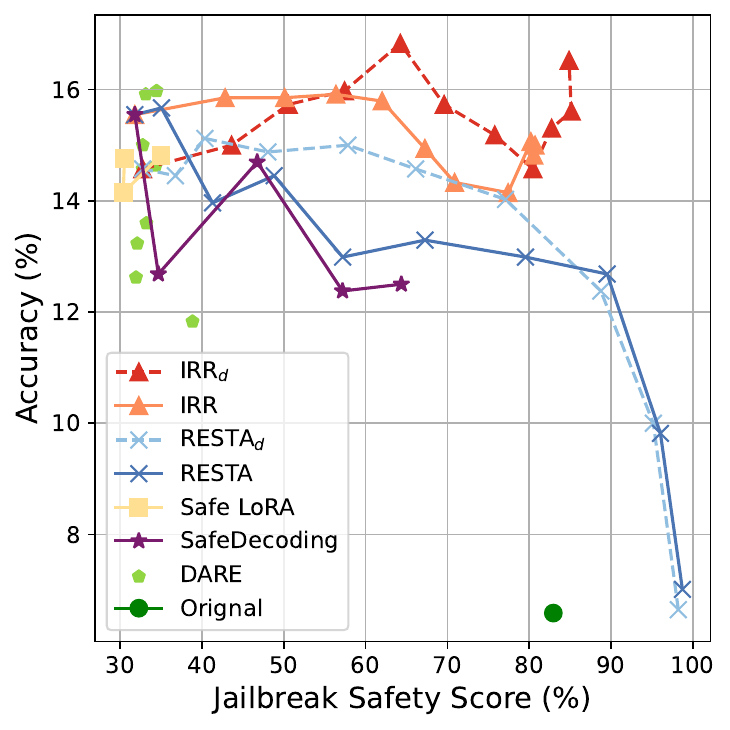}
		\caption{Code}
		\label{LoRA_code_jailbreak}
	\end{subfigure}
	\centering
	\begin{subfigure}{0.3\textwidth}
		\centering
		\includegraphics[width=1.0\textwidth]{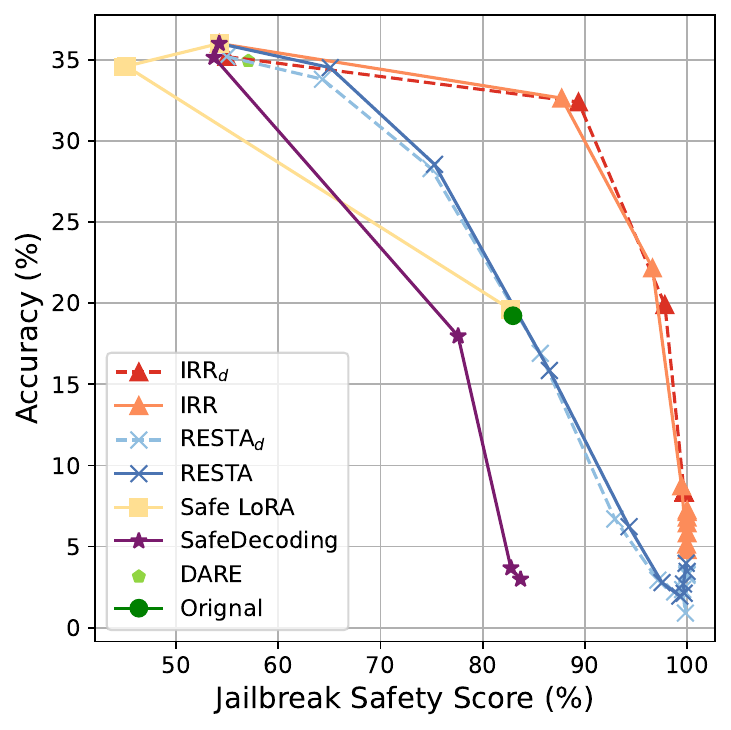}
		\caption{Chinese}
		\label{LoRA_chinese_jailbreak}
	\end{subfigure}
	\caption{LoRA experimental results. We present the trends of "task performance versus safety score" on the \textbf{Jailbreak attacks}. Our method, IRR, outperforms baseline methods by improving safety while maintaining strong downstream task performance.}
	\label{fig:LoRA_jailbreak_performance_pareto}
\end{figure*}

\section{LoRA fine-tuning Experimental}
\label{sec:appendix_lora_experimental_details}
We also conducted experiments on SFT models fine-tuned with LoRA \citep{hulora} and evaluated their safety using the harmful benchmark and jailbreak attack. The specific experimental results are shown in Figure \ref{fig:LoRA_safety_performance_pareto} and Figure \ref{fig:LoRA_jailbreak_performance_pareto}. We found that for LoRA fine-tuning, the IRR method is also effective, achieving Pareto improvements in both safety enhancement and maintaining downstream task performance.

\section{Multilingual Safety}
\label{sec:appendix_multilingual}
\begin{figure*}[t]
	\centering
	\begin{subfigure}{0.3\textwidth}
		\centering
		\includegraphics[width=1.0\textwidth]{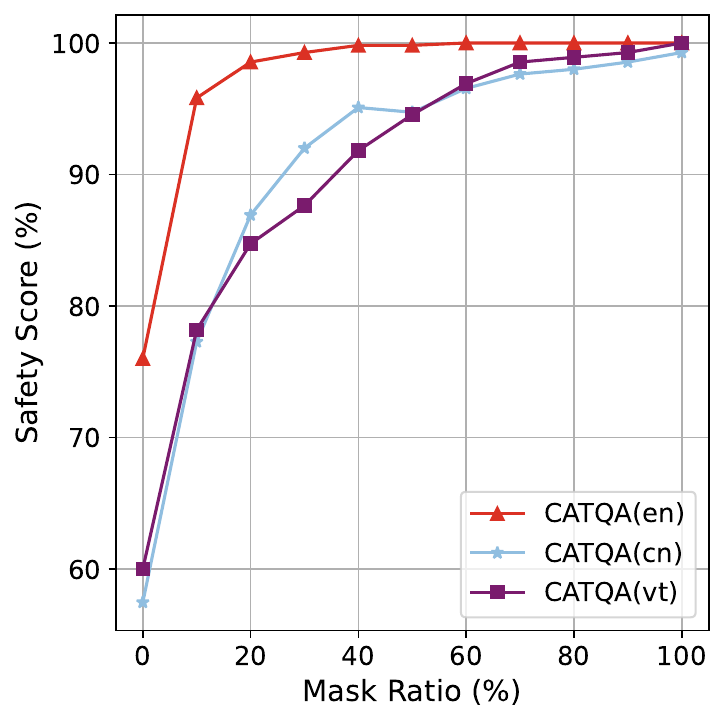}
		\caption{Code}
		\label{code_cross_language_jailbreak}
	\end{subfigure}
	\centering
	\begin{subfigure}{0.3\textwidth}
		\centering
		\includegraphics[width=1.0\textwidth]{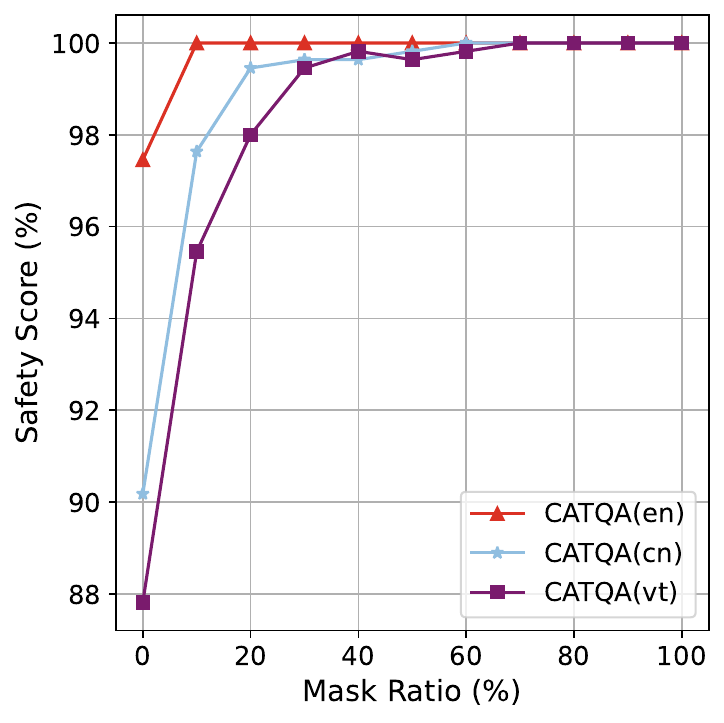}
		\caption{Chinese}
		\label{chinese_cross_language_jailbreak}
	\end{subfigure}
	\caption{We conduct a safety evaluation on the English, Chinese, and Vietnamese versions of CATQA. We compare the safety changes of harmful queries across different mask ratios and languages.}
	\label{fig:Code_Chinese_multilingual_catqa}
\end{figure*}

We assessed the safety of SFT models fine-tuned on the Code (CodeAlpaca-20k) and Chinese (Alpaca-Chinese) datasets using the English, Chinese, and Vietnamese versions of the harmful benchmark CATQA (see Figure \ref{fig:Code_Chinese_multilingual_catqa}). As shown in Figure~\ref{fig:Code_Chinese_multilingual_catqa}, the safety scores across different languages improve as the masking ratio increases for both the Code and Chinese datasets.

\section{Ablation Experiments}
\label{sec:appendix_ablations}
\begin{figure*}[t]
	\centering
	\begin{subfigure}{0.32\textwidth}
		\centering
		\includegraphics[width=1.0\textwidth]{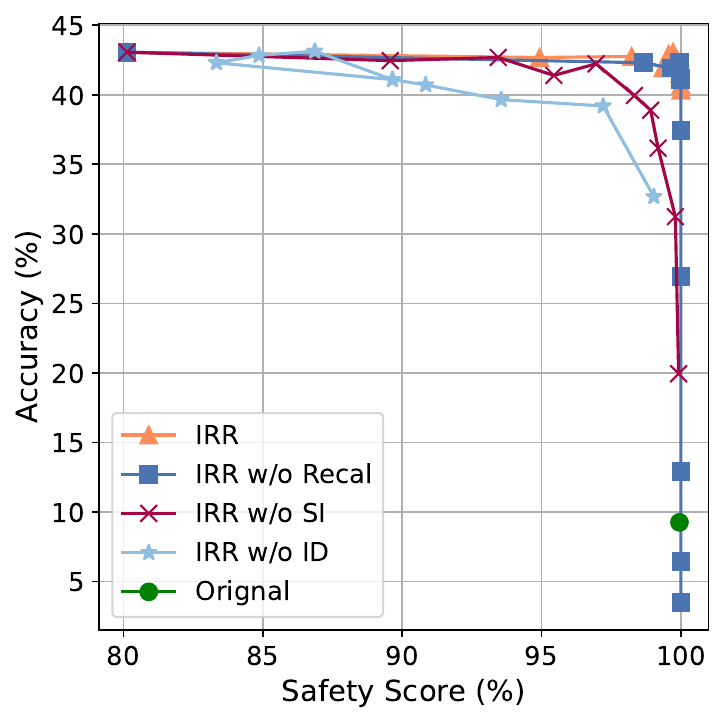}
		\caption{Math}
		\label{Math_Safety}
	\end{subfigure}
	\centering
	\begin{subfigure}{0.32\textwidth}
		\centering
		\includegraphics[width=1.0\textwidth]{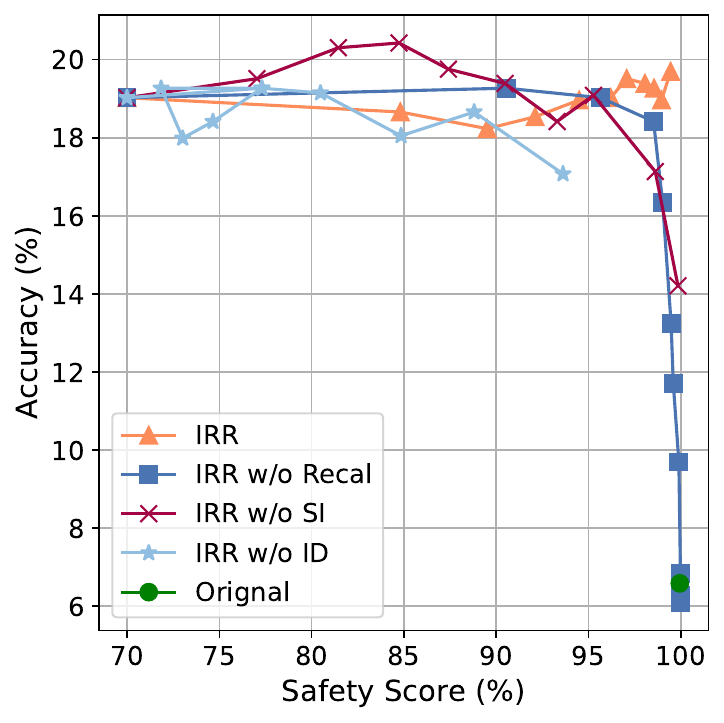}
		\caption{Code}
		\label{Code_Safety}
	\end{subfigure}
	\centering
	\begin{subfigure}{0.32\textwidth}
		\centering
		\includegraphics[width=1.0\textwidth]{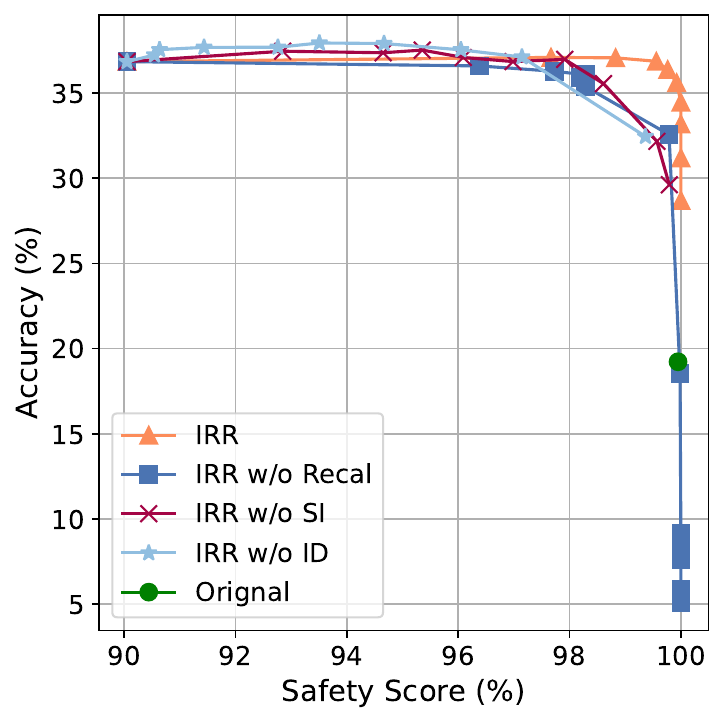}
		\caption{Chinese}
		\label{chinese_Safety}
	\end{subfigure}
	\caption{We present the results of the IRR ablation study using “task performance vs. safety” curves on harmful benchmarks. Our method effectively identifies unsafe delta parameters and, combined with the calibration step, successfully preserves downstream task performance.}
	\label{fig:safety_ablation_analysis}
\end{figure*}

We conducted an ablation study on the IRR method and presented the trade-off curve between downstream task performance and safety on the harmful benchmark in Figure \ref{fig:safety_ablation_analysis}.

\section{Details about Llama-3-8B-Instruct experiment}
\label{sec:appendix_llama3_experimental_details}
\begin{figure}[!ht]
    \centering
    \begin{minipage}{0.7\linewidth}
        \centering
        \includegraphics[width=\linewidth]{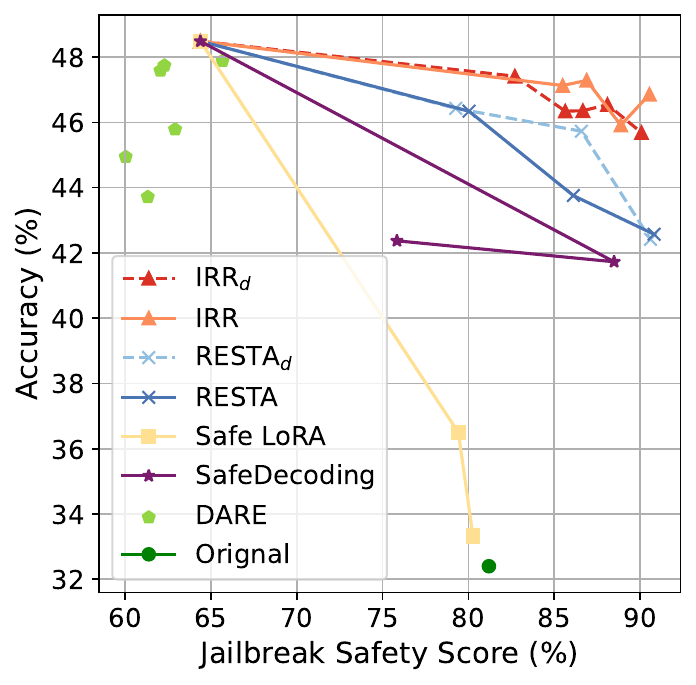}
    \end{minipage}\hfill
    \caption{We fine-tune the Llama-3-8B-Instruct model using LoRA on the MathInstruct dataset. We then evaluate the impact of IRR and baseline methods on the safety and mathematical capabilities of the SFT model.}
    \label{fig:appendix_llama3_mathinstruct_exp_jailbreak}
\end{figure}

In the Llama3 experiments, we fine-tuned Meta-Llama-3-8B-Instruct \citep{dubey2024llama} using LoRA. The query and value matrices in LoRA were tuned with a rank of \(r=8\).  The training batch size was set to 64, and the learning rate configured as \(2.5 \times 10^{-4}\). Fine-tuning was performed on the MathInstruct dataset \citep{yuemammoth}. We set the mask ratio in the IRR method to 1\%, 2\%, 3\%, 4\%, and 5\%.

To evaluate the mathematical capabilities of the fine-tuned model, we conducted few-shot evaluations on GSM8K \citep{cobbe2021training}, Math \citep{hendrycks2measuring}, AQuA \citep{ling-etal-2017-program}, simuleq \citep{koncel-kedziorski-etal-2016-mawps}, numglue \citep{mishra-etal-2022-numglue}, MMLU STEM \citep{hendrycksmeasuring}, and SAT math \citep{zhong-etal-2024-agieval} datasets. The evaluation was implemented using the math-evaluation-harness framework \footnote{\url{https://github.com/ZubinGou/math-evaluation-harness}}.

Meta-Llama-3-8B-Instruct does not include a default system prompt, so no system prompt is added during training or inference. Both the training and inference of Llama-3 use the chat template.


\begin{figure}[!ht]
    \centering
    \begin{minipage}{0.7\linewidth}
        \centering
        \includegraphics[width=\linewidth]{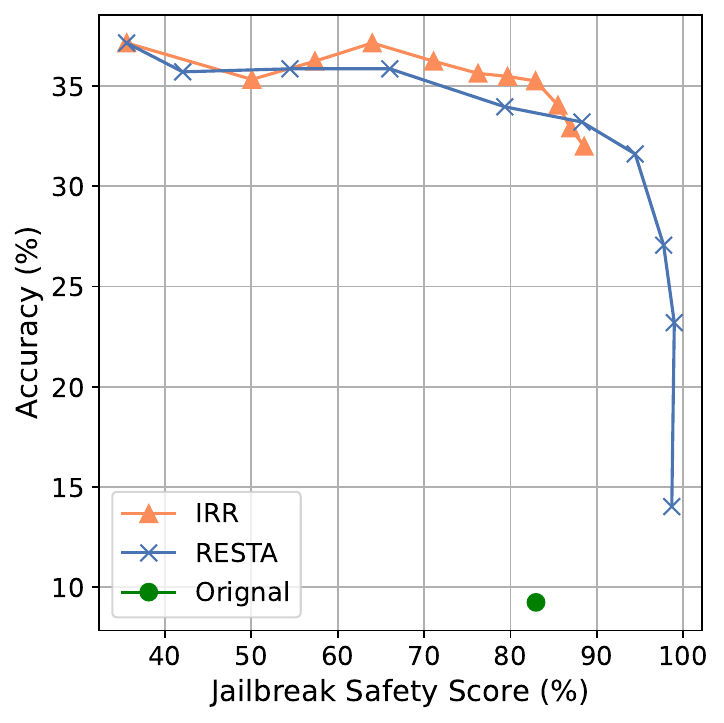}
    \end{minipage}\hfill
    \caption{Results of IRR and RESTA using the new safety vector.}
    \label{fig:appendix_IRR_with_new_safety_vector}
\end{figure}

\section{IRR with New Safety Vector}
\label{sec:appendix_IRR_new_safety_vector}
We report the performance of the IRR method using the new safety vector in a jailbreak scenario. The results in Figure \ref{fig:appendix_IRR_with_new_safety_vector} show that the IRR method achieves better safety and downstream task performance in defending against jailbreak attacks.

\section{Computational Complexity of IRR}
\label{sec:Computational_Complexity_of_IRR}
To implement IRR, we leverage a computationally efficient technique called SparseGPT \citep{frantar2023sparsegpt} to compute the inverse Hessian matrix, which is a critical component of the OBS computation. The computational complexity of calculating the inverse Hessian as described in SparseGPT can be divided into three main parts: 

\textbf{Initial Hessian Calculation.} The time complexity for calculating the initial Hessian matrix is \(O(nd^2_{\text{col}})\), where \(n\) represents the number of input samples and \(d_{\text{col}}\) is the number of columns in the matrix.

\textbf{Hessian Inversion Iteration.} The iterative inversion of the Hessian matrix has a time complexity of \(O(d^3_{\text{col}})\), which remains manageable even for large models.

\textbf{Reconstruction Process.} The pruning or reconstruction process based on the inverse Hessian involves a complexity of \(O(d^3_{\text{col}} + d^2_{\text{row}}d_{\text{col}})\), where \(d_{\text{row}}\) denotes the number of rows in the matrix. This ensures that the process is computationally feasible even for models with a large number of parameters.

In summary, considering the hidden dimension \(d_{\text{hidden}}\) of Transformer models, the overall computational complexity aligns with \(O(d^3_{\text{hidden}})\). This indicates a significant improvement in efficiency compared to exact reconstruction methods, demonstrating that our approach is computationally practical even for very large models.

\begin{table*}[!ht]
    \centering
    \resizebox{0.8\textwidth}{!}{ 
    \begin{tabular}{c|cccc}
        \hline
        \textbf{Data Size} & \textbf{Time (s)} & \textbf{Acc (\%)} & \textbf{Safety Score (\%)} & \textbf{Jailbreak Safety Score (\%)} \\
        \hline
        8 & 227.03 & 42.43 & 99.26 & 85.52 \\
        64 & 260.29 & 42.02 & 99.12 & 83.74 \\
        128 & 298.23 & 42.14 & 99.20 & 83.41 \\
        1,000 & 839.22 & 42.06 & 99.16 & 82.45 \\
        \hline
    \end{tabular}
    }
    \caption{Results of running IRR on the GSM8K full fine-tuned models}
    \label{tab:IRR_Time_Full_Tuned_Model}
\end{table*}

\begin{table*}[!ht]
    \centering
    \resizebox{0.8\textwidth}{!}{ 
    \begin{tabular}{c|cccc}
        \hline
        \textbf{Data Size} & \textbf{Time (s)} & \textbf{Acc (\%)} & \textbf{Safety Score (\%)} & \textbf{Jailbreak Safety Score (\%)} \\
        \hline
        8 & 57.10 & 35.81 & 99.51 & 81.60 \\
        64 & 67.57 & 35.66 & 99.53 & 81.13 \\
        128 & 79.34 & 35.79 & 99.52 & 81.01 \\
        1,000 & 242.97 & 35.53 & 99.52 & 80.29 \\
        \hline
    \end{tabular}
    }
    \caption{Results of running IRR on the GSM8K LoRA fine-tuned models}
    \label{tab:IRR_Time_LoRA_Model}
\end{table*}

\begin{table}[!ht]
    \centering
    \resizebox{0.7\linewidth}{!}{ 
    \begin{tabular}{c|c}
        \hline
        \textbf{Model Size} & \textbf{Time (s) / Layer}  \\
        \hline
        7B & 8.13  \\
        13B & 10.12  \\ 
        70B & 31.08  \\
        \hline
    \end{tabular}
    }
    \caption{Results of running IRR on the GSM8K LoRA fine-tuned models}
    \label{tab:IRR_Time_Per_Layer}
\end{table}

\section{Time Consumption of IRR} 
To further investigate the time efficiency of the IRR method, we analyzed the time required to run IRR. Specifically, we measured the time needed to execute IRR on both full and LoRA fine-tuned models and estimated the time required when scaling to larger models. The testing platform used was an A100 80GB GPU.

\subsection{Time for the IRR Method}
We used the GSM8K \citep{cobbe2021training} dataset to measure the time required for the IRR method on full and LoRA fine-tuned Llama-2-7b-chat models \citep{touvron2023llama}. The IRR method requires a specific amount of downstream task data to compute the Hessian matrix. We used 8, 64, 128, and 1000 samples from downstream tasks. The mask ratio started at 10\% and increased incrementally by 10\% up to 100\%. We reported the average results for time consumption, downstream task accuracy, safety scores, and jailbreak safety scores.

As shown in Table \ref{tab:IRR_Time_Full_Tuned_Model} and Table \ref{tab:IRR_Time_LoRA_Model}, using a smaller amount of calibration data can reduce the time required for the IRR method without significantly affecting downstream task performance or safety scores. This demonstrates that the IRR method can achieve faster safety realignment with less calibration data. For example, when utilizing the IRR method on a LoRA fine-tuned 7B model with 64 calibration samples, only 67.57 seconds are needed. In comparison, fine-tuning the LoRA model on the GSM8K dataset for 5 epochs takes at least 1.55 hours. Furthermore, the IRR method allows for safety realignment without increasing inference time, highlighting its advantages.

\subsection{Estimating Time Efficiency for Scaling to Larger Models}
Next, we will estimate the running time of the IRR method when it is applied to larger models. We used 64 samples from downstream tasks to calculate the Hessian matrix and measured the time needed for the IRR method to run on a single layer of the model.

According to the Table \ref{tab:IRR_Time_Per_Layer}, for larger models, such as the full fine-tuned 40-layer Llama2-13B, the IRR method takes about 404.8 seconds to complete. For the 80-layer 70B model, it takes approximately 2,486.4 seconds.

\section{Limited GPU Memory Resources}

In scenarios with limited GPU memory resources, the IRR method, based on SparseGPT, computes the inverse Hessian matrix by traversing each layer of the model. After calculating the Hessian matrix for the current layer, it proceeds to the next layer's computation. This approach theoretically allows for only the current layer to be loaded into GPU memory during the calculations, thereby enabling safe realignment within the constraints of limited GPU memory resources.

\end{document}